\documentclass[11pt, letterpaper]{cmu}

\usepackage[all]{hypcap}
\usepackage[comma,numbers,sort,compress]{natbib}
\usepackage{hyperref}

\hypersetup{
    colorlinks = true,
    citecolor = {magenta},
}

\usepackage{microtype}
\usepackage{graphicx}
\usepackage{subfigure}
\usepackage{booktabs}
\usepackage{float}
\usepackage{tabularx}
\usepackage{amsmath}
\usepackage{amssymb}
\usepackage{mathtools}
\usepackage{amsthm}
\usepackage{mathrsfs}
\usepackage{nicefrac}
\usepackage{dsfont}
\usepackage{enumitem}
\usepackage{float}

\usepackage{xspace}
\usepackage[capitalize,noabbrev]{cleveref}
\usepackage{subcaption}
\usepackage{wrapfig}
\usepackage{lipsum}
\usepackage{listings}

\usepackage{amsmath}
\usepackage{amssymb}
\usepackage{mathtools}
\usepackage{amsthm}
\usepackage{bbm}
\usepackage{fleqn, tabularx}
\usepackage{multirow}
\usepackage[export]{adjustbox}
\usepackage{colortbl}

\usepackage{algpseudocode}
\usepackage{setspace}

\usepackage{color}
\definecolor{deepblue}{rgb}{0,0,0.5}
\definecolor{deepred}{rgb}{0.6,0,0}
\definecolor{deepgreen}{rgb}{0,0.5,0}
\definecolor{boost_correct_to_correct}{HTML}{66C2A5}
\definecolor{default_correct_to_correct}{HTML}{fc8d62}
\definecolor{dup_correct_to_correct}{HTML}{8da0cb}
\definecolor{new_correct_to_correct}{HTML}{e78ac3}

\definecolor{MyBlue}{RGB}{32,102,214}

\definecolor{MyRed}{RGB}{200,60,50}

\usepackage[capitalize,noabbrev]{cleveref}

\definecolor{lightblue}{rgb}{0.22,0.45,0.70}
\newcommand{\highlight}[1]{\textbf{\textit{\textcolor{lightblue}{#1}}}}

\usepackage[most]{tcolorbox}

\newtcolorbox{tipbox}{
  enhanced,
  boxrule=0.6pt,
  left=6pt,right=6pt,top=6pt,bottom=6pt,
  colback=blue!3,
  colframe=blue!40!black,
  arc=3pt,
}

\newcommand\pythonstyle{\lstset{
basicstyle=\ttfamily\footnotesize,
language=Python,
morekeywords={self, clip, exp, mse_loss, uniform_sample, concatenate, logsumexp},              
keywordstyle=\color{deepblue},
emph={MyClass,__init__},          
emphstyle=\color{deepblute},   
stringstyle=\color{deepgreen},
frame=single,                       
showstringspaces=false
}}

\lstnewenvironment{python}[1][]
{
\pythonstyle
\lstset{#1}
}
{}

\newcommand\pythoninline[1]{{\pythonstyle\lstinline!#1!}}

\definecolor{blanchedalmond}{rgb}{1.0, 0.92, 0.8}
\definecolor{carmine}{rgb}{0.59, 0.0, 0.09}
\definecolor{lightblue}{rgb}{0.22,0.45,0.70}

\renewcommand{\mathbf}{\boldsymbol}

\makeatletter
\def\Ddots{\mathinner{\mkern1mu\raise\p@
\vbox{\kern7\p@\hbox{.}}\mkern2mu
\raise4\p@\hbox{.}\mkern2mu\raise7\p@\hbox{.}\mkern1mu}}
\makeatother

\numberwithin{equation}{section}

\definecolor{amaranth}{rgb}{0.9, 0.17, 0.31}
\definecolor{antiquebrass}{rgb}{0.8, 0.58, 0.46}
\definecolor{antiquefuchsia}{rgb}{0.57, 0.36, 0.51}
\definecolor{chromeyellow}{rgb}{0.31, 0.47, 0.26}

\usepackage[dvipsnames]{xcolor}
\definecolor{maj5}{HTML}{2b8cbe}
\definecolor{maj5Imp}{HTML}{084081}

\definecolor{seq5wo}{HTML}{d95f0e}
\definecolor{seq5woImp}{HTML}{662506}

\definecolor{seq5w}{HTML}{6a51a3}
\definecolor{seq5wImp}{HTML}{3f007d}

\definecolor{selfwo}{HTML}{d95f0e}
\definecolor{selfwoImp}{HTML}{662506}

\definecolor{selfw}{HTML}{6a51a3}
\definecolor{selfwImp}{HTML}{3f007d}

\definecolor{glorewo}{HTML}{d95f0e}
\definecolor{glorewoImp}{HTML}{662506}

\definecolor{glorew}{HTML}{6a51a3}
\definecolor{glorewImp}{HTML}{3f007d}

\definecolor{vstar}{HTML}{d95f0e}
\definecolor{vstarImp}{HTML}{662506}

\makeatletter
\def\mathcolor#1#{\@mathcolor{#1}}
\def\@mathcolor#1#2#3{%
  \protect\leavevmode
  \begingroup
    \color#1{#2}#3%
  \endgroup
}
\makeatother

\usepackage[textsize=tiny]{todonotes}
\usepackage{algorithm}
\usepackage{amssymb}

\Crefformat{equation}{#2Eq.\;(#1)#3}

\Crefformat{figure}{#2Figure #1#3}
\Crefformat{assumption}{#2Assumption #1#3}
\Crefname{assumption}{Assumption}{Assumptions}

\usepackage{crossreftools}
\usepackage[suppress]{color-edits}

\usepackage{dsfont}
\usepackage{nicefrac}

\newtcolorbox{analysisbox}[1][]{
    enhanced jigsaw,
    colback=white,
    colframe=blue!75!black,
    fonttitle=\bfseries,
    boxsep=5pt,
    left=5pt,
    right=5pt,
    top=5pt,
    bottom=5pt,
    title=#1,
}

\tcbset{
  aibox/.style={
    width=\linewidth,
    top=8pt,
    bottom=4pt,
    colback=blue!6!white,
    colframe=black,
    colbacktitle=black,
    enhanced,
    center,
    attach boxed title to top left={yshift=-0.1in,xshift=0.15in},
    boxed title style={boxrule=0pt,colframe=white,},
  }
}
\newtcolorbox{AIbox}[2][]{aibox,title=#2,#1}
\definecolor{lightblue}{rgb}{0.22,0.45,0.70}

\usepackage{pifont}
\usepackage{soul}
\definecolor{highlightmistake}{RGB}{255, 179, 179}
\definecolor{highlightcorrect}{RGB}{179, 255, 179}

\title{IsoCompute Playbook: Optimally Scaling Sampling Compute for LLM RL}
\author{
Zhoujun~Cheng$^{\dagger,\ddagger,*}$, 
Yutao~Xie$^{\dagger,*}$, 
Yuxiao~Qu$^{\S,*}$, 
Amrith~Setlur$^{\S,*}$, 
Shibo~Hao$^{\dagger,\ddagger}$, 
Varad~Pimpalkhute$^{\ddagger}$,   
Tongtong~Liang$^{\dagger}$, 
Feng~Yao$^{\dagger}$, 
Zhengzhong~Liu$^{\ddagger}$, 
Eric~Xing$^{\ddagger,\S}$, 
Virginia~Smith$^{\S}$, 
Ruslan~Salakhutdinov$^{\S}$, 
Zhiting~Hu$^{\dagger}$, 
Taylor~Killian$^{\ddagger}$, 
Aviral~Kumar$^{\S}$ \\
\vspace{0.5em}
$^{\dagger}$UC San Diego \,\, 
$^{\ddagger}$MBZUAI-IFM \,\, 
$^{\S}$Carnegie Mellon University \\
$^{*}$Equal contribution \,\,  Website: \href{https://compute-optimal-rl-llm-scaling.github.io/}{\texttt{https://compute-optimal-rl-llm-scaling.github.io/}}
}
\date{}

\correspondingauthor{z6cheng@ucsd.edu, yux076@ucsd.edu, yuxiaoq@andrew.cmu.edu, asetlur@andrew.cmu.edu}

\begin{abstract}

\end{abstract}

\begin{document}

\maketitle

\begin{figure}[h]
    \centering
    \vspace{-1.0cm}
    \includegraphics[width=0.95\linewidth]{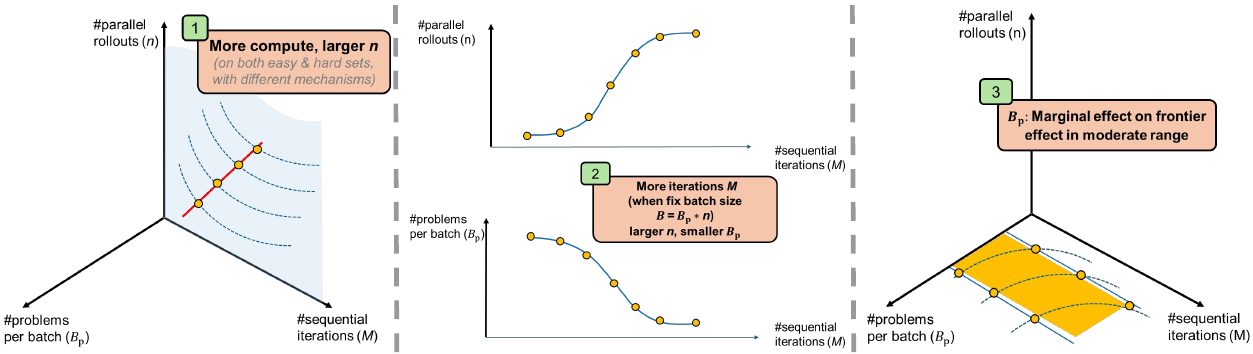}
    \caption{\footnotesize{\textbf{ Compute-optimal sampling for LLM RL.} 
We study allocation of sampling compute along three axes: parallel rollouts per problem ($n$), problems per batch ($B_{\text{p}}$), and sequential iterations ($M$), where the total compute is
$C = B_{\text{p}} \cdot n \cdot M$.
We find that: (1) optimal number of rollouts $n$ increases with the compute budget $C$; (2) easy and hard problem sets exhibit similar scaling trends but arise from different underlying mechanisms; (3) under a constraint on $B = B_{\text{p}} \cdot n$, the optimal strategy prioritizes larger $B_{\text{p}}$ (smaller $n$) at low compute budgets, and shifts toward larger $n$ (smaller $B_{\text{p}}$) at high compute budgets to maximize performance; and (4) $B_{\text{p}}$ has only a marginal effect on performance when kept within a moderate range.
\vspace{-2mm}
}}
    \label{fig:main_fig}
\end{figure}

{\absfont \textbf{Abstract:} While scaling laws guide compute allocation for LLM pre-training, analogous prescriptions for reinforcement learning (RL) post-training of large language models (LLMs) remain poorly understood. We study the compute-optimal allocation of sampling compute for on-policy RL methods in LLMs, framing scaling as a compute-constrained optimization over three resources: parallel rollouts per problem, number of problems per batch, and number of update steps. We find that the compute-optimal number of parallel rollouts per problem increases predictably with compute budget and then saturates. This trend holds across both easy and hard problems, though driven by different mechanisms: solution sharpening on easy problems and coverage expansion on hard problems. We further show that increasing the number of parallel rollouts mitigates interference across problems, while the number of problems per batch primarily affects training stability and can be chosen within a broad range. Validated across base models and data distributions, our results recast RL scaling laws as prescriptive allocation rules and provide practical guidance for compute-efficient LLM RL post-training.}

\vspace{-0.3cm}
\section{Introduction}
\label{sec:introduction}
\vspace{-0.2cm}

A blocker in scaling up reinforcement learning (RL) for large language models (LLMs) is the absence of a \emph{concrete workflow}: a recipe that tells practitioners \emph{what} to scale, \emph{how} to scale it, and \emph{what outcomes of scaling} one should expect. In many areas of modern AI, such workflows are enabled by empirical scaling laws~\cite{arxiv-org-1712-00409,arxiv-org-2001-08361,arxiv-org-2203-15556}, where initial experiments reveal predictable relationships between performance and resources (e.g., compute, data). These laws guide compute allocation, model selection, and hyperparameter choices. In this paper, our goal is to understand and build analogous scaling laws for RL post-training of LLMs.

In contrast to pre-training or supervised learning, scaling behavior in RL is far less understood due to the tight coupling between \textbf{exploration} (data collection) and \textbf{optimization} (learning from data). Recent work has begun to characterize scaling behavior in classical deep RL~\citep{arxiv-org-2104-03113,arxiv-org-2301-13442,value-scaling-github-io-value-scaling-github-io, rybkin2025valuebaseddeeprlscales, mccandlish2018empirical}.

However, in the LLM setting, this line of study remains in its infancy. The most relevant prior results show that, under a given fixed problem mixture, RL reward curves exhibit clean sigmoidal behavior when trained for longer~\cite{arxiv-org-2510-13786}, or that RL performance scales with model size in a manner reminiscent of pre-training~\cite{arxiv-org-2509-25300, rybkin2025valuebaseddeeprlscales, value-scaling-github-io-value-scaling-github-io}. While informative, these results stop short of addressing the central question that often plagues practitioners running RL: \textbf{\emph{how to allocate resources when setting up an RL run for a base model?}} Given a base model, a problem distribution, and a fixed compute budget, how should one spend this compute to maximize downstream performance?

We address a big part of this question in this work by studying the optimal allocation of \textbf{sampling compute} in LLM RL. To this end, we conduct a series of experiments across three base models (Qwen2.5-7B-Instruct, Qwen3-4B-Instruct, and Llama~3.1-8B-Instruct), covering diverse training configurations and problem distributions, including easy, hard, and skewed mixtures of prompts (also referred to as problems). Concretely, we operate in a setting where we optimize some binary notion of success or reward on a mixture of problems. Our analysis reveals a nuanced picture of scaling. Unlike pre-training, scaling behavior in RL is governed not only by total compute, but also by the interaction between the base model and the prompt distribution. Nevertheless, under \emph{healthy} and stable training recipes, we are able to derive \emph{predictable} allocation rules for key hyperparameters in LLM RL as a function of sampling compute for a base model. Concretely, for on-policy RL methods that optimize LLM policies using multiple parallel rollouts per sequential gradient step, we make the following observations as in Figure~\ref{fig:main_fig}, validated across about \textbf{$120,000$ H200-hours} of RL experiments on top of three base models.

In short, our findings are as follows. \textbf{\textit{First}}, the compute-optimal number of parallel rollouts per input problem increases with the sampling compute budget and then saturates. This means that as more compute becomes available, performance improves by allocating more rollouts per problem rather than simply training longer.
\textbf{\textit{Second}}, this scaling trend holds across both easy and hard problem sets, but for different reasons. On easy problems, increasing the number of rollouts primarily sharpens performance on already solvable prompts, reflected in improvements in worst@k metrics. On hard problems, larger numbers of rollouts are essential for discovering rare successful trajectories, leading to gains in best@k and improved coverage. \textbf{\textit{Third}}, under fixed hardware constraints (e.g., a fixed number of GPUs), performance is relatively insensitive to the number of unique problems per batch compared to the number of rollouts per problem. This suggests a simple allocation strategy: prioritize sampling more problems when the compute budget admits only a small number of sequential training steps, and shift toward more rollouts per problem as the number of training steps increases. On hard problems, this trade-off is more nuanced and depend on the evaluation metric.
\textbf{\textit{Finally}}, while these scaling trends generalize across base models and datasets, the absolute value of the compute-optimal number of rollouts is context-dependent and saturates at different points depending on model capacity, dataset size, and problem difficulty.

\vspace{-0.2cm}
\section{Problem Statement}
\label{sec:problem_statement}
\vspace{-0.1cm}

We consider post-training an LLM using binary outcome-reward RL on a fixed dataset of problems. We focus on rollout-based on-policy algorithms such as GRPO~\citep{arxiv-org-2402-03300-2}, which generate multiple rollouts per prompt and optimize the policy using group-normalized advantages. Concretely, for each prompt, we sample $n$ rollouts, score them with a 0/1 outcome reward, and compute advantages by centering (i.e., subtracting mean) and normalizing (i.e., dividing by standard deviation) rewards \emph{within} this group.

Unlike classical RL, where data acquisition costs arise from interacting with an external simulator, RL for LLMs in single-turn settings typically generates its own training data during optimization. As a result, the primary resource constraint is \textbf{\emph{sampling compute}}, which is proportional to the total number of generated rollouts, denoted by $C$. We divide this budget into three parts: \textbf{(1)} \emph{problem batch size} ($B_\text{p}$), the number of unique prompts sampled per step; and \textbf{(2)} \emph{group size} ($n$), the number of parallel rollouts generated per problem in a single update; \textbf{(3)} \emph{update iterations} ($M$), the number of sequential gradient updates. $M$ governs the amount of \textbf{\emph{sequential}} compute, while $B_\text{p}$ and $n$ govern the amount of \textbf{\emph{parallel}} compute. The effective batch size per iteration is $B = B_\text{p} \cdot n$, and the total sampling compute factorizes as:
\begin{align*}
    C = B_\text{p} \cdot n \cdot M.
\end{align*}

\textbf{Formalizing the goal of our study.}
Let $\mathcal{A}(B_\text{p}, n, M)$ denote an RL algorithm instantiated with these hyperparameters, and let $\mathcal{P}(\cdot)$ denote a scalar performance metric of the resulting model (e.g., reward or pass rate). Rather than treating our goal as exactly solving a single constrained optimization problem, we study the following scaling questions under a fixed sampling budget $C_0$:
\begin{align}
\label{eq:scaling_opt}
    (B^*_\text{p}(C_0), n^*(C_0), M^*(C_0))
    \in \arg\max_{B_\text{p}, n, M} \;
    \mathcal{P}\!\left(\mathcal{A}(B_\text{p}, n, M)\right)
    \quad
    \text{s.t.}
    \quad
    B_\text{p} \cdot n \cdot M \le C_0.
\end{align}

Specifically, we ask: \emph{(i)} how performance varies as sampling compute is allocated across $B_\text{p}$, $n$, and $M$; and \emph{(ii)} how the optimal allocation changes as the budget $C_0$ increases.

\textbf{Predictable scaling laws.}
In this work, we say a scaling law is \emph{predictable} if the dependence of performance and optimal allocation on compute budget follows a stable trend that can be well-approximated from measurements at smaller budgets and then extrapolated to larger budgets. Concretely, our aim is to characterize how $\mathcal{P}$ and the induced optimum $(B^*_\text{p}(C_0), n^*(C_0), M^*(C_0))$ vary with $C_0$, and whether these trends admit simple functional forms that support the prediction of compute-optimal allocation.

\vspace{-0.2cm}
\section{Designing a Healthy RL Recipe}
\vspace{-0.1cm}

Predictable scaling trends emerge from Equation~\ref{eq:scaling_opt} only if the performance of the algorithm $\mathcal{P}(\mathcal{A}(B_\text{p}, n, M))$ varies smoothly with respect to changes in $B_\text{p}, n, M$ under the constraint on compute $B_\mathrm{p} \cdot n \cdot M \leq C_0$.

\begin{wrapfigure}{r}{0.48\textwidth}
  \vspace{-0.1cm}
  \centering
  \includegraphics[width=\linewidth]{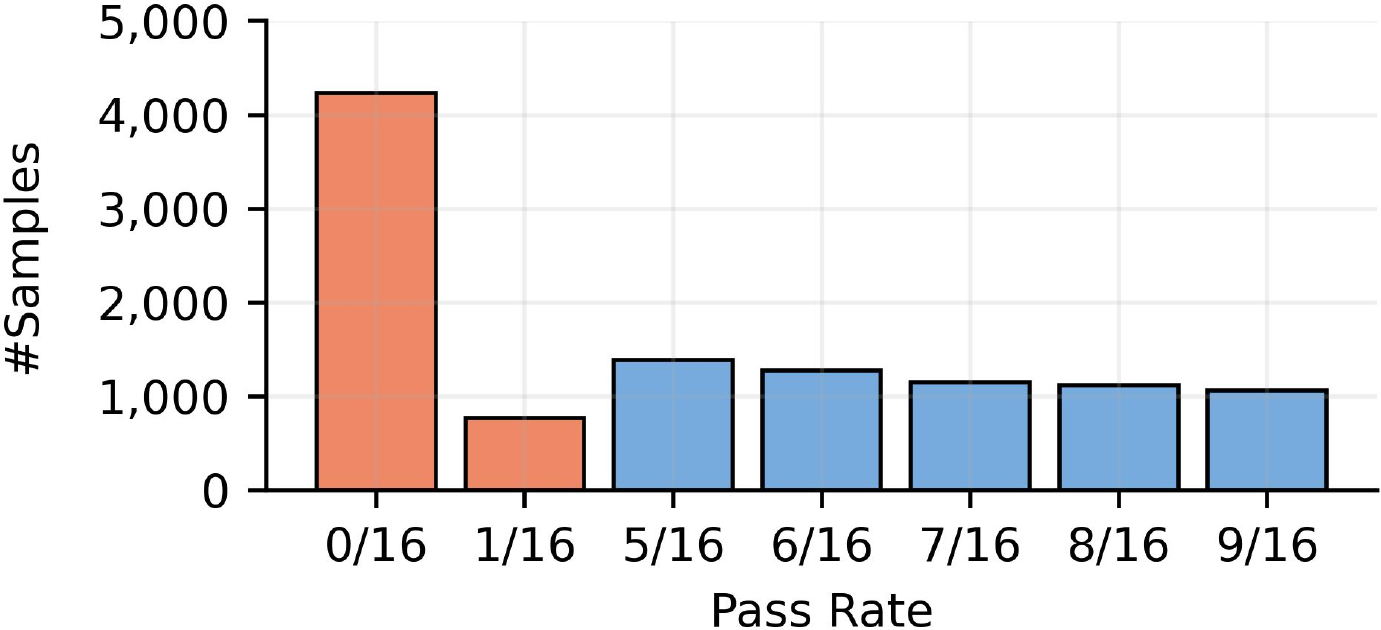}
  \vspace{-0.45cm}
  \caption{\footnotesize{\textbf{Difficulty distribution of Easy vs. Hard problems.}
  We split problems into Easy and Hard sets according to pass@16 (average pass rate over 16 generations per problem).}}
  \label{fig:sec2_data_dist}
  \vspace{-0.5cm}
\end{wrapfigure}

A core desideratum, therefore, is that the RL algorithm $\mathcal{A}$ exhibits stable training dynamics as sampling compute is scaled. In practice, na\"ive implementations often violate this requirement~\citep{liu2025prorl}. 

Because hyperparameters such as $(B_\text{p}, n, M)$ jointly control both data collection and optimization, changing them without care \emph{can} induce instabilities in training, making performance highly non-smooth and obscuring underlying scaling structure. Therefore, before studying scaling laws, we first establish a ``healthy'' RL recipe whose dynamics remain stable across a range of sampling compute budgets. We find that in our setting, training stability is most consistently governed by three factors: \textbf{(i)} problem difficulty relative to the base model, \textbf{(ii)} use of entropy and KL regularization, and \textbf{(iii)} learning-rate scaling with the effective batch size ($B = B_\text{p} \cdot n$).

\textbf{Factor 1: Dataset difficulty distribution.} We find that the difficulty of a problem relative to the base model~\citep{snell2024scalingllmtesttimecompute} strongly affects stability of an RL run. On easy prompts where the base model already samples correct rollouts frequently, RL can quickly drive down entropy and collapse exploration~\citep{arxiv-org-2505-22617};
on hard prompts, reward is rarely observed and optimization instead demands more exploration. We quantify difficulty by \textit{avg@16}, the base model’s average accuracy over 16 rollouts (Qwen2.5-7B-Instruct), which measures the {ease of experiencing reward} during RL rather than human difficulty. Hence, we construct difficulty-based splits from the Guru-Math dataset~\cite{arxiv-org-2506-14965}, each with 300 in-domain validation samples: \textbf{(a) Easy}, with \textit{avg@16} $\in [0.3, 0.6]$ (6k samples), and \textbf{(b) Hard}, with \textit{avg@16} $\in [0.0, 0.0625]$ (5k samples). These datasets will be used for our main experiments (Figure~\ref{fig:sec2_data_dist}).

\begin{wrapfigure}{r}{0.65\textwidth}
  \centering
  \vspace{-0.1cm}
  \includegraphics[width=\linewidth]{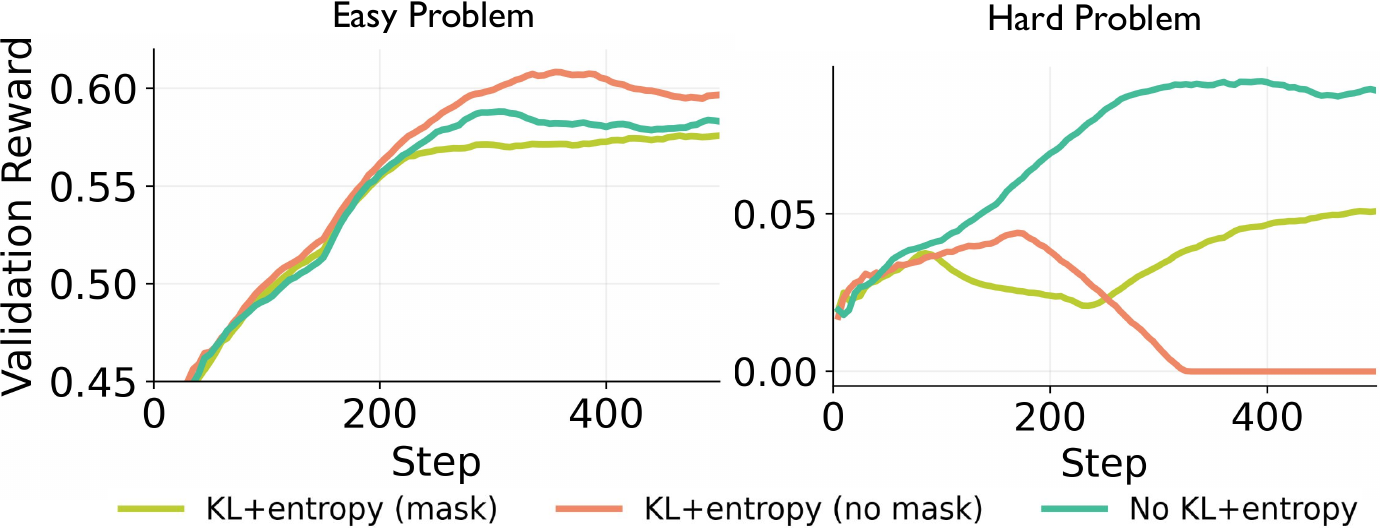}
  \caption{\footnotesize{\textbf{Regularization ablations on Easy and Hard.}
  On the Easy set, standard KL+Entropy regularization achieves the best reward.
  On the Hard set, these regularizers destabilize training even with zero-variance filtering; disabling them yields significantly more stable optimization and higher reward.}}
  \label{fig:sec2_kl_ent_ablation}
  \vspace{-3.0mm}
\end{wrapfigure}
\textbf{Factor 2: Entropy and KL-divergence regularization.} Problem difficulty manifests clearly in token-level entropy and, more weakly, in the KL divergence to the base model (Figure~\ref{fig:sec2_kl_ent_ablation}), both of which serve as sensitive indicators of optimization health. Token-level entropy governs the degree of exploration during generation, while the KL term anchors the policy and limits excessive drift from the base model~\citep{yu2025dapo}. On easy problems, insufficient entropy regularization often leads to premature entropy collapse, causing optimization to stall. In contrast, on hard problems, entropy regularization alone can trigger entropy and response-length explosion, as policy gradients aggressively push toward rare successful trajectories~\citep{qu2026popelearningreasonhard}. In this regime, a KL term can be effective at delaying or preventing early-stage instability, although it is typically unnecessary if training is stable. Hence, whenever we employ an entropy bonus, we pair it with a KL anchor. While applying zero-variance filtering~\citep{arxiv-org-2510-13786} to these terms mitigates instability, we find it suboptimal in performance. In our experiments, we apply both KL and entropy regularization on easy problem sets, where collapse is the dominant failure mode, and remove both on hard problem sets to avoid instability. Importantly, our scaling results are robust to this choice of regularization, provided that training remains stable.

\begin{figure}[t]
\centering

\begin{minipage}{0.55\textwidth}
  \centering
  \vspace{-0.1cm}
  \includegraphics[width=\linewidth]{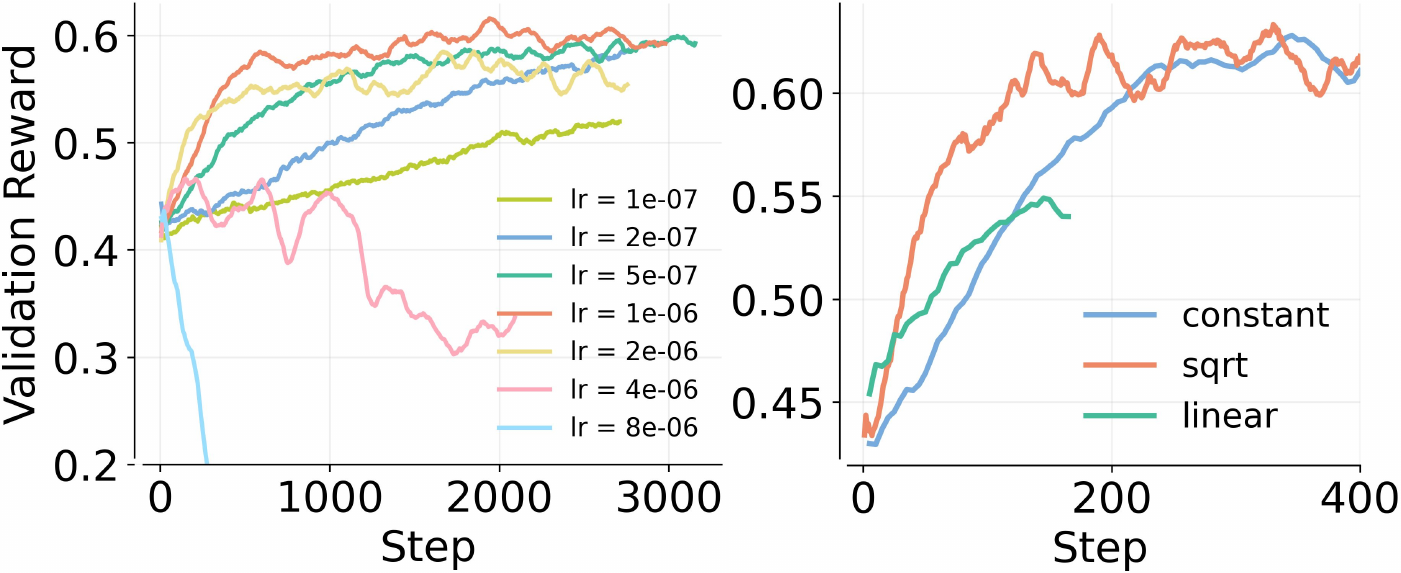}
  \vspace{-0.6cm}
  \caption{\footnotesize{\textbf{LR scaling strategy.} Square-root scaling ($\sqrt{B}$) outperforms linear and constant scaling at large batch sizes ($B=8192$).}}
  \label{fig:lr_and_config}
\end{minipage}
~\vline~
\begin{minipage}{0.40\textwidth}
  \centering
  \captionsetup{justification=centering}
  \captionof{table}{\footnotesize{\textbf{Final recipe.} Details of the final recipe used in our study.}}
  \label{tab:recipe_summary}
  \resizebox{0.88\linewidth}{!}{\begin{tabular}{lcc}
  \toprule
  \textbf{Hyperparameter} & \textbf{Easy} & \textbf{Hard} \\
  \midrule
  KL Regularization      & Yes & No \\
  Entropy Regularization & Yes & No \\
  Zero-var Filter        & No  & No \\
  LR Scaling             & $\sqrt{B}$ & $\sqrt{B}$ \\
  \bottomrule
  \end{tabular}}
\end{minipage}

\vspace{-0.4cm}
\end{figure}

\textbf{Factor 3: Learning rate scaling.} Since we vary batch size ($B$) significantly in our scaling laws study, we require a robust LR scaling rule. We first identify a base learning rate $\eta_{\text{base}} = 10^{-6}$ at $B=1,024$~(Figure~\ref{fig:lr_and_config} (left)). Similar to~\citep{yang2022tensorprogramsvtuning}, we then compare constant, linear, and square-root scaling strategies. As shown in Figure~\ref{fig:lr_and_config} (right), \textbf{square-root scaling} ($\eta \propto \sqrt{B}$) provides the best trade-off, enabling faster convergence than using a constant learning rate while avoiding the instability of linear scaling.
Based on these findings, we adopt the configuration listed in the Table~\ref{tab:recipe_summary} for the main experiments. See Appendix~\ref{app:exp_detail} for full experiment details that we study in this paper.

\begin{AIbox}{Key Takeaways: Designing a Healthy RL Recipe}
\begin{enumerate}[itemsep=2pt]
  \item RL training exhibits distinct behaviors depending on problem difficulty. We therefore explicitly \textbf{curate and control for both Easy and Hard datasets} to ensure the recipe is robust to different saturation points and exploration requirements. On heterogeneous datasets discussed later, we use the recipe corresponding to the Hard dataset to avoid instability.

  \item The necessity of regularization changes with difficulty. \textbf{Easy tasks} benefit from KL divergence and entropy constraints to prevent premature collapse, whereas \textbf{Hard tasks} achieve peak performance when these loss terms are removed to enable stable training. Training on mixed datasets is most stable when no KL divergence or entropy are used.

  \item Do set learning rates as a function of the total batch size $B$. Among the schemes we compared, the \textbf{square-root learning-rate scaling} strategy performs best.
\end{enumerate}
\end{AIbox}

\vspace{-0.2cm}
\section{Allocating Sampling Compute Optimally}
\label{sec:main_results}
\vspace{-0.1cm}

\begin{wrapfigure}{r}{0.45\textwidth}
  \centering
  \vspace{-0.5cm}
    \includegraphics[width=0.95\linewidth]{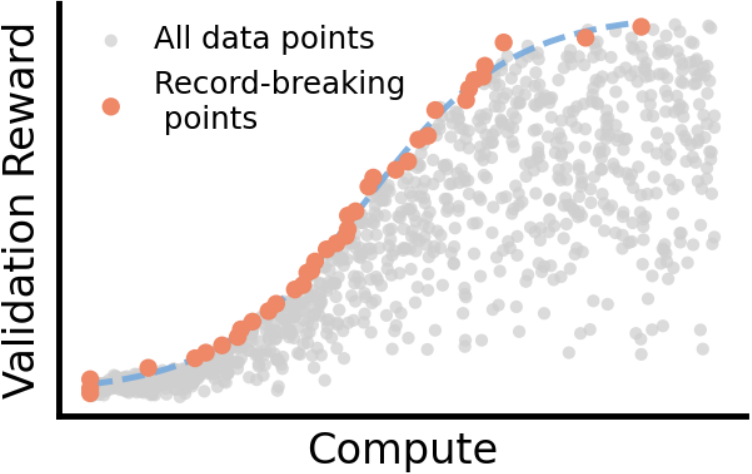}
    \vspace{-0.3cm}
\caption{\footnotesize{\textbf{Illustration of record-breaking points.}
Gray dots show validation reward points from multiple training runs, while orange dots mark record-breaking points, defined as the earliest (smallest compute) points that enter a higher discretized reward bin than all previous points. The dashed curve shows the monotonic fit over the retained points on the performance frontier.}}
    \label{fig:frontier_cartoon}
    \vspace{-0.35cm}
\end{wrapfigure}

We now present empirical results that address our central question: \emph{{given a fixed sampling compute budget, how should it be allocated across RL sampling dimensions to maximize performance?}} Recall that the total sampling compute scales as $C$ $\propto B_\text{p} \cdot {n \cdot M}$. To study allocation strategies, we sweep over values of $(B_\text{p}, n, M)$ across a range of budgets $C$. For a fixed compute budget $C = C_0$, we evaluate multiple allocations and define the \textbf{\textit{compute-optimal frontier}} as the highest i.i.d. validation set reward achievable using total compute $C_0$. Repeating this procedure for increasing values of $C_0$ yields a family of frontiers that characterize how optimal allocation evolves with available compute.

\textcolor{lightblue}{\textbf{Data analysis workflow.}}
To derive our scaling law fits, we subsample each training run to a compact set of
\textbf{\textit{record-breaking}} points along the learning curve, defined by validation reward as a function of increasing compute. A record-breaking point is the earliest step at which the validation reward exceeds all previously observed values; Figure~\ref{fig:frontier_cartoon} provides an illustration. To robustly identify such improvements, we select the first step at which the discretized reward enters a higher bin. We restrict attention to record-breaking points because non-record-breaking checkpoints are dominated by earlier checkpoints from the same run that achieve equal or better validation reward with less compute, and thus cannot lie on the compute-optimal frontier. Including all checkpoints would overweight long, highly correlated training trajectories and bias the fit toward suboptimal intermediate points rather than the best-achievable performance envelope.

We then fit a monotonic function to these record-breaking points to obtain prescriptions for the optimal values of $n$, $B_\text{p}$, and $M$. Because this preprocessing preserves the ordering of points along the compute axis, it does not introduce spurious non-monotonicity and yields a faithful estimate of the performance frontier (see Appendix~\ref{app:exp_detail}, Figure~\ref{fig:frontier_tutorial}, for an illustration).
 
\textcolor{lightblue}{\textbf{Experimental setup.}} We sweep over valid configurations $(B_\text{p}, n)$, where $B_\text{p} \in \{2^5, \dots, 2^{10}\}$ and $n \in \{2^3, \dots, 2^{11}\}$, using uniform intervals on a log scale. Due to parallelism limits of the available GPUs, we additionally incorporate a hardware-driven batch size constraint $B_\text{p} \cdot n \leq B_{\max}$. We set $B_{\max} = 65{,}536$ for the Easy set and $16{,}384$ for the Hard set. For each run, the number of update steps $M$ increases as training proceeds. We use a smaller value of $B_{\max}$ for the Hard set to allow for more sequential iterations within a fixed total compute budget. See Appendix~\ref{app:exp_detail} for full details regarding the experimental setup. 
We adopt \emph{rollouts} rather than \emph{tokens} as our metric of compute, since the number of generated tokens that the model will produce during RL training cannot be reliably estimated \emph{a priori} and thus provides limited guidance for compute allocation. That said, we show in Appendix~\ref{app:token_view} that translating our scaling trends to measure compute in terms of tokens still yields similar conclusions regarding allocation rules in practice.

We study compute-optimal allocation rules in three settings that isolate distinct resource trade-offs: \textbf{(1)} $n$ vs.\ $M$ (parallel rollouts vs. sequential updates); \textbf{(2)} $n$ vs.\ $B_{\text{p}}$ (parallel rollouts vs. number of problems per batch); and \textbf{(3)} joint allocation across all resources. Each setting corresponds to a practical scenario in which a practitioner must allocate limited compute across competing dimensions.

\vspace{-0.2cm}
\subsection{Parallel Samples $n$ vs Sequential Iterations $M$}
\label{sec:q1}
\vspace{-0.1cm}
In this section, we fix the number of problems $B_\text{p}$ and study {the trade-off between parallel samples $n$ and sequential iterations $M$} under a fixed budget $C$.

\begin{wrapfigure}{r}{0.63\textwidth}
  \centering
    \includegraphics[width=\linewidth]{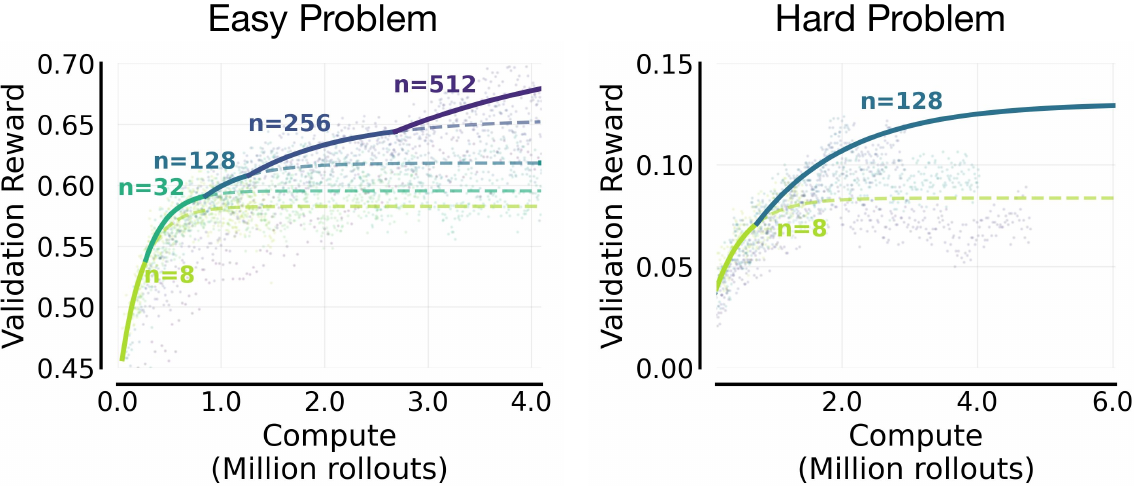}
    \vspace{-0.65cm}
    \caption{\footnotesize{\textbf{Validation reward vs. compute $(B_\text{p}=32)$}. The frontier shifts to larger $n$ as compute increases. For easy problems \textit{\textbf{(left)}}, large $n$ dominates at high compute where small $n$ plateaus. Hard problems \textit{\textbf{(right)}} show the same trend but saturate earlier with a smaller $n$.}}
    \label{fig:sec3_q1_fixBprob_frontier}
    \vspace{-0.4cm}
\end{wrapfigure}

\begin{wrapfigure}{r}{0.63\textwidth}
  \vspace{-0.2cm}
  \centering
    \includegraphics[width=\linewidth]{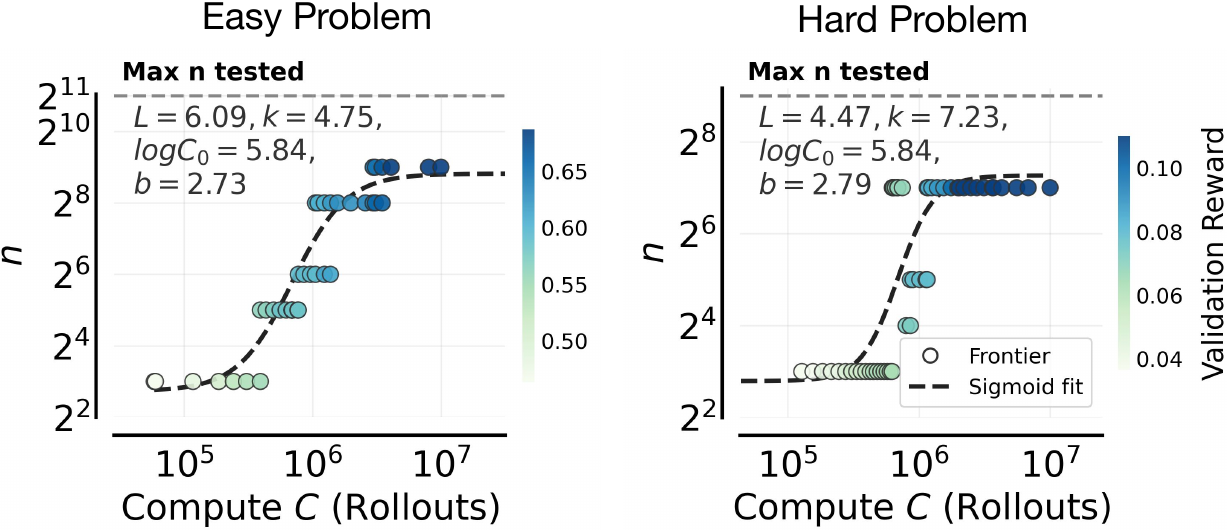}
    \vspace{-0.65cm}
    \caption{\label{fig:sec3_q1_fixBprob_sigmoid} \footnotesize{\textbf{Compute-optimal scaling of parallel rollouts $n$} ($B_\text{p}=32$). The optimal value of rollouts $n$ \textbf{shifts systematically higher} as the total sampling compute increases. Points show a running-average estimate of the frontier-attaining  $n^*(C)$ at each compute budget (colored by reward), and the red curves fit a sigmoid parameterizing $\log n$ as a function of $\log C$.}}
    \vspace{-0.35cm}
\end{wrapfigure}

{\textbf{Fitting workflow.}} We plot reward vs compute $C$ and fit a \emph{\textbf{monotonic sigmoid}} to summarize how the validation set reward (avg@4) scales with compute for that $n$. As mentioned above, we then define the \emph{compute-optimal frontier} as the upper envelope of these fitted curves (see Figure~\ref{fig:sec3_q1_fixBprob_frontier}). Then, to indicate which $n$ lies on the frontier at each compute level, we color the frontier by $n^*(C)$, which is the value of $n$ whose fitted compute–reward curve achieves the compute-optimal frontier \textbf{up to} $C$. Finally, in Figure~\ref{fig:sec3_q1_fixBprob_sigmoid}, we fit a log-log plot to show  $n^*(C)$ as a function of  $C$ to summarize the empirical scaling behavior. We make four important observations in this setting.

\textcolor{MyBlue}{{1) The value of $n$ lying on the compute-optimal frontier shifts higher as the sampling compute $C$ increases (Figure~\ref{fig:sec3_q1_fixBprob_frontier}).}} It is natural to expect larger values of $n$ to be generally favorable at higher compute budgets, analogous to prior work~\citep{arxiv-org-2510-01180}, since increasing $n$ lowers policy-gradient variance but it requires more sampling compute. Consistent with this belief, the frontier-attaining $n^*(C)$ shifts to larger values as $C$ grows, and we observe the same trend on both the Easy and Hard problem sets. Smaller values of $n$ exhibit rapid initial gains but plateau at a relatively lower compute regime, whereas larger $n$ sustain improvement over a broader compute range. \emph{\textbf{This behavior also suggests that parallel and sequential compute are not interchangeable.}} Choosing $n$ so that we are able to perform sufficient sequential updates $M$ is necessary to achieve strong performance.

\textcolor{MyBlue}{{2) Compute-optimal values of $n$ are well-approximated by a sigmoid function of $C$ (Figure~\ref{fig:sec3_q1_fixBprob_sigmoid}).}} We next aim to fit a functional relationship for the compute optimal value $n^*(C)$ as a function of the available compute $C$. A natural first step is to hypothesize an appropriate functional form. As shown in Figure~\ref{fig:sec3_q1_fixBprob_sigmoid}, increasing $C$ admits larger compute optimal values of $n$, and over a substantial range this relationship appears approximately linear on a log-log scale. The key question is whether this growth continues indefinitely or eventually saturates. Empirically, we observe a clear saturation. Even when evaluating rollout values up to $n=2,048$, values significantly larger than the saturation point, they fail to extend the frontier, with $n=512$ continuing to dominate.

\textbf{We argue that saturation is expected} when training a fixed base model and a fixed problem set. To build intuition as to why, it is perhaps helpful to view increasing $n$ as analogous to spending more compute per gradient step. In supervised learning, increasing capacity alone does not reduce validation error beyond a certain point unless additional training data is available. This principle also underlies pre-training scaling rules from Chinchilla~\cite{arxiv-org-2203-15556} that prescribe scaling both pre-training data and model capacity together. Perhaps most closely related to the RL training setup in this study, \citet{arxiv-org-2406-14532} shows that increasing $n$  cannot overcome limitations imposed by a fixed problem set for rejection fine-tuning. As a result, the compute optimal value of $n$ must eventually saturate even for RL, as we observe. We validate this hypothesis regarding a fixed data size in Section \ref{sec:bigger_picture}, where we show how the saturation point shifts given {a different base model, problem set size, and distribution}.

\textcolor{MyBlue}{{3) Next, we find that the compute-optimal allocation \textit{trend} remains consistent across difficulty levels, although we find harder sets prefer smaller values of $n$ (Figure~\ref{fig:sec3_q1_fixBprob_sigmoid})}}. We find that the compute optimal allocation trend remains consistent across problem difficulty. On both problem sets, the compute optimal value of $n$ increases with total compute $C$ before eventually plateauing. However, the plateau occurs clearly at {{smaller}} values of $n$ on harder problems. In particular, very large values of $n$, such as $n=512$, yield lower final performance on the hard set and do not lie on the compute optimal frontier. \textbf{\textit{This suggests that task difficulty imposes an upper bound on how large $n$ can be used effectively}}. While it may seem intuitive that harder problems should benefit from larger $n$ due to increased sampling right away, we observe the opposite behavior in practice. On sufficiently hard problem sets, increasing $n$ allocates substantial compute to problems where the model receives little or no learning signal. In contrast, smaller values of $n$ focus optimization on the subset of prompts where nonzero signal is already present and meaningful improvement is possible. Therefore, it is better to use a smaller value of $n$ to increase the frequency of parameter updates (small $n$, large $M$, more epochs on the same subset of problems) that exploits reachable gains, rather than spending larger $n$ on problems that are persistently unsolved.

\begin{wrapfigure}{r}{0.6\textwidth}
  \vspace{-0.65cm}
  \centering
    \includegraphics[width=\linewidth]{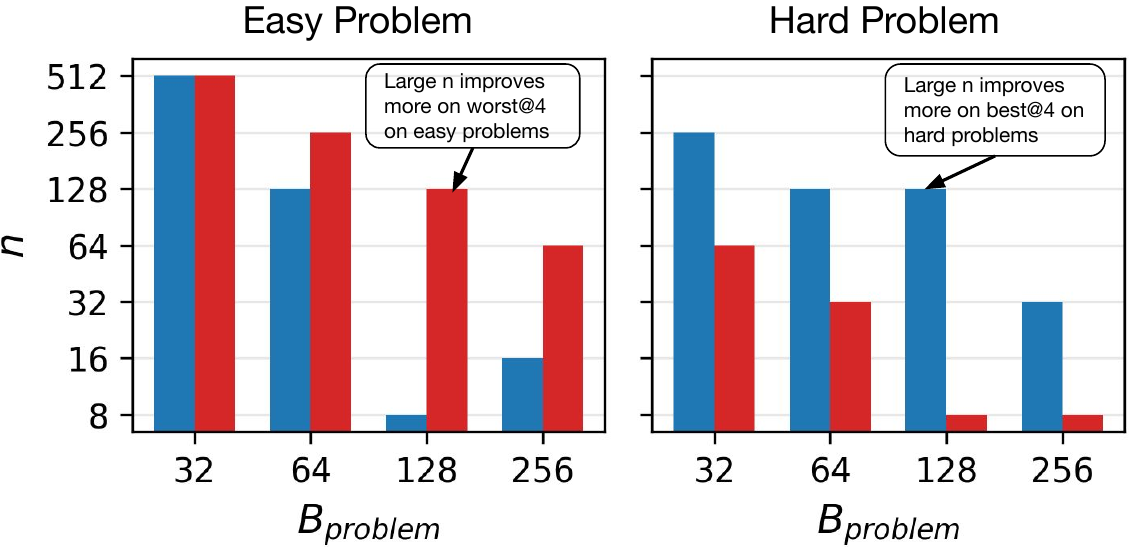}
    \vspace{-0.75cm}
    \caption{\footnotesize{\textbf{Different mechanisms of how $n$ values optimize \highlight{best@4} vs. \highlight{worst@4} on easy and hard problems.} Bars show the $n$ maximizing reward for a given $B_\text{p}$. On the Easy set~\textbf{\textit{(left)}}, the optimal $n$ for best@4 is smaller than for worst@4, indicating that improving robustness requires more parallel rollouts than for coverage. Conversely, on the Hard set~\textbf{\textit{(right)}}, a larger $n$ is needed to improve best@4, while worst@4 saturates at smaller $n$.}}
    \label{fig:sec3_q1_fixBprob_bestworstk}
    \vspace{-0.2cm}
\end{wrapfigure}

\textcolor{MyBlue}{{4) Optimization dynamics on the easy and hard sets and the role of various performance metrics (Figure~\ref{fig:sec3_q1_fixBprob_bestworstk}).}} We saw above that a smaller value of $n$ was more preferable for optimizing validation \textit{average reward} (avg@4 per problem) and attributed this to solving new problems vs. solving the same problems, but better. We now aim to better understand these optimization dynamics and evaluate how $n^*(C)$ changes if we were to change \emph{the target performance metric} we study. In particular, we consider two metrics: \textbf{\textit{best@k}} (or pass@k), defined as the fraction of problems where {at least one} response out of $k$ is correct, which measures the model's \textbf{coverage} over problems; and \textbf{\textit{worst@k}}, defined as the fraction of problems where all $k$ responses are correct, which we examine to measure the degree to which we can ``\textbf{sharpen}'' around the right solution (i.e., robustness).

Modulo compute-optimality, a larger value of $n$ coupled with as many sequential update steps as needed, should in principle, result in higher values for both \textit{best@k} and \textit{worst@k} on a training dataset. However, this is not quite the case when compute is bounded. We empirically identify the optimal values of $n^*(C)$ for obtaining the highest \textit{best@k} and \textit{worst@k} scores on the validation set, across different $B_\mathrm{p}$ values for the largest value of $C$, and show this number in Figure~\ref{fig:sec3_q1_fixBprob_bestworstk}. We choose $k=4 \ll n$ we study, so that none of the trends in Figure~\ref{fig:sec3_q1_fixBprob_bestworstk} are ``edge'' cases or artifacts of fitting/statistical error. Surprisingly, we now see an interesting divergence in trends on the Easy and Hard sets.

\textbf{Results.} On the easy set, {a larger $n$ is compute-optimal for \textit{worst@4} (sharpening) performance, whereas smaller values of $n$ are compute-optimal for the \textit{best@4} performance.} This means that a larger $n$ primarily improves by sharpening more on easy problems, while a smaller $n$ suffices to sample one correct rollout (expected since the set is easy). Conversely, for hard problems, a larger $n$ is more critical for pushing up \textit{best@4} (coverage), while a relatively smaller $n$ is compute-optimal for \textit{worst@4} (sharpening). However, there is a limit beyond which a larger $n$ does not improve coverage on new problems in a compute-optimal way: optimal values here are generally lower than on the easy set. On the \textit{Extremely Hard} set consisting of all pass@128 = 0 problems (Appendix~\ref{app:other_results}; Figure~\ref{fig:appx_passall0}), we see a clearer tradeoff of coverage and sharpening: while larger $n$ improves \textit{best@k}, it degrades \textit{worst@k} and lowers average reward. When targeting average reward, the optimal $n$ on hard problems is the value that balances coverage and sharpening well. These results imply that the target metric itself dictates the landscape of compute-optimal $n$.

\begin{AIbox}{Key Takeaways}
\begin{enumerate}
  \item The compute-optimal $n$ frontier {shifts systematically higher} as the total sampling compute increases, and is well fit by \textbf{a sigmoid curve} for all datasets.
  \item The source of gains from large $n$ {shifts with training data difficulty}: scaling $n$ improves \emph{sharpening} (worst@4) on the Easy set, but expands \emph{coverage} (best@4) on the Hard set.
\end{enumerate}
\end{AIbox}

\begin{AIbox}{Workflow Prescriptions}
\begin{itemize}
  \item Depending on the composition of the problem set and how effectively the base model can learn from it, the mechanism driving performance improvements may differ. We recommend diagnosing the \emph{mode} of improvement for your base model on the prompt set (e.g., sharpening vs.\ coverage), and using this to set $n$ as a function of the available compute budget $C$.
\end{itemize}
\end{AIbox}

\vspace{-0.2cm}
\subsection{Bounded Batch Compute: Trading off $B_\text{p}$ with $n$}
\label{sec:q2}
\vspace{-0.1cm}

Next, we study a different setup, where we wish to allocate a fixed total batch size $B$ into the number of prompts used and the number of rollouts per prompt used. This question is important in practical settings where hardware parallelism (e.g., number of GPUs or data-parallel) is fixed, and a practitioner needs to make this compute allocation. In such cases, $B$ is often chosen as the largest rollout batch size that saturates sampling throughput ("system batch size"). We additionally experimented with $B_\text{p}={8}$ and $16$ for the Easy set under fixed $B$ to locate the upper and lower bounds for values of $B_\text{p}$ and $n$.

We specify the number of sequential iterations $M$ \textbf{\textit{a priori}} and seek allocations of $B_\text{p}$ and $n$ under a fixed total batch budget $B_\text{p} \cdot n \leq B$ that maximize performance. We observe the following:

\begin{wrapfigure}{r}{0.6\textwidth}
  \centering
  \vspace{-0.4cm}
  \includegraphics[width=\linewidth]{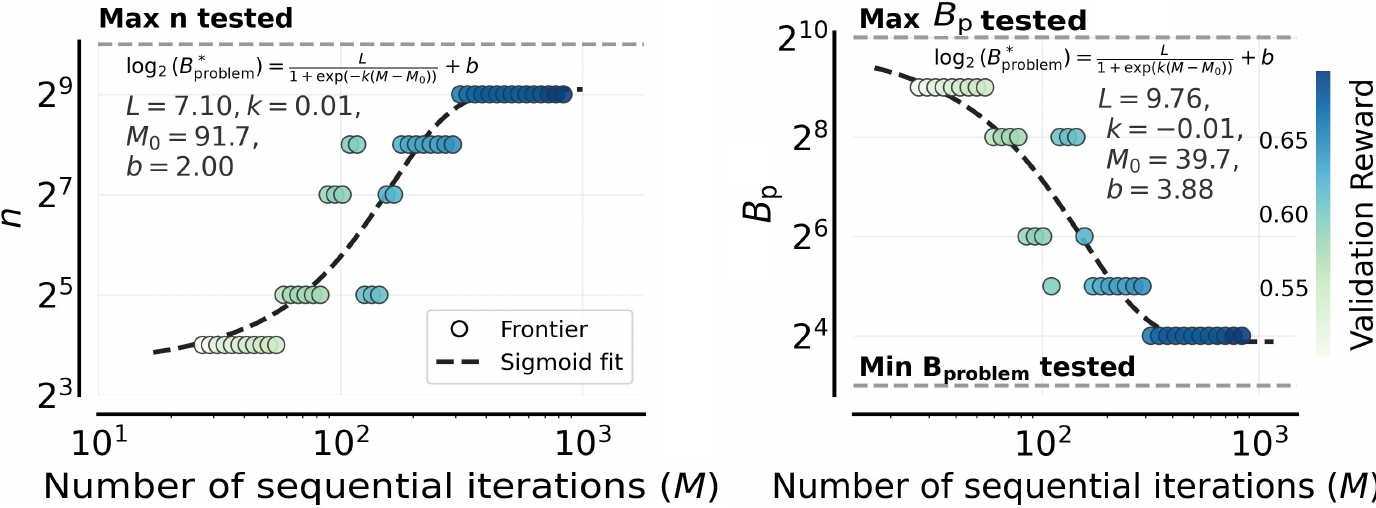}
  \vspace{-0.65cm}
\caption{\footnotesize{Compute-optimal allocation shifts from $B_{\text{p}}$ to $n$ under a fixed total batch size constraint on easy set.}}
  \label{fig:sec3_q2_fixB_sigmoid}
  \vspace{-0.3cm}
\end{wrapfigure}

\textcolor{MyBlue}{{1) On the easy problems, allocate more parallel compute $n$ when sequential steps $M$ is large (Figure~\ref{fig:sec3_q2_fixB_sigmoid}).}} In this regime, we examine the compute-optimal value of $n$ under a fixed total batch size (illustrated with $B=8,192$ only in Figure~\ref{fig:sec3_q2_fixB_sigmoid}), as $M$ varies. The optimal choice $n^*(M)$ exhibits a sigmoidal dependence on $M$. This behavior suggests that when more sequential updates are available, it is preferable to allocate additional compute toward increasing $n$, rather than increasing $B_\text{p}$. The corresponding compute-optimal number of prompts $B_\text{p}^*(M)$ decreases with the sampling compute according to an (inverse) sigmoid. In contrast, when $M$ is small, allocating batch size toward a larger $B_\text{p}$ is more effective, as it enables many more epochs of training within a given total sequential updates. On the Hard set, however, the scaling behavior is less consistent. The compute-optimal value $n^*(M)$ exhibits a non-monotonic dependence on $M$ (see Appendix~\ref{app:other_results}, Figure~\ref{fig:appx_fixB_easy}-\ref{fig:appx_fixB_hard}), which implies a similarly irregular trend for the optimal $B_\text{p}$. \emph{\textbf{This is one of the differences we see across Easy and Hard sets.}}

\textcolor{MyBlue}{{2) Why do we observe different trends on the Easy and Hard sets in this setup?}} 
As discussed previously, reward can be increased either by scaling $n$, which improves the quality of signal obtained per problem, or by scaling  $B_\text{p}$, which broadens the set of problems used for training. On the Easy set, where the base model already produces correct rollouts with high probability, the dominant bottleneck is sample quality, making larger values of $n$ preferable as $M$ increases.
On the Hard set, however, the optimal allocation depends strongly on the \emph{stage} of training. When the number of sequential updates $M$ is small, low values of $n$ are ineffective at extracting gradient signal, even if training is restricted to a subset of problems. As $M$ increases and the model begins to receive signal on a limited set of problems, increasing $B_\text{p}$ becomes preferable, as it prevents overfitting to this small subset. Finally, at larger values of $M$, once training has stabilized across a set of problems, it becomes possible to increase $n$ again without sacrificing coverage, and the compute-optimal allocation shifts back toward larger $n$.

\begin{wrapfigure}{r}{0.6\textwidth}
  \centering
  \vspace{-0.6cm}
  \includegraphics[width=0.97\linewidth]{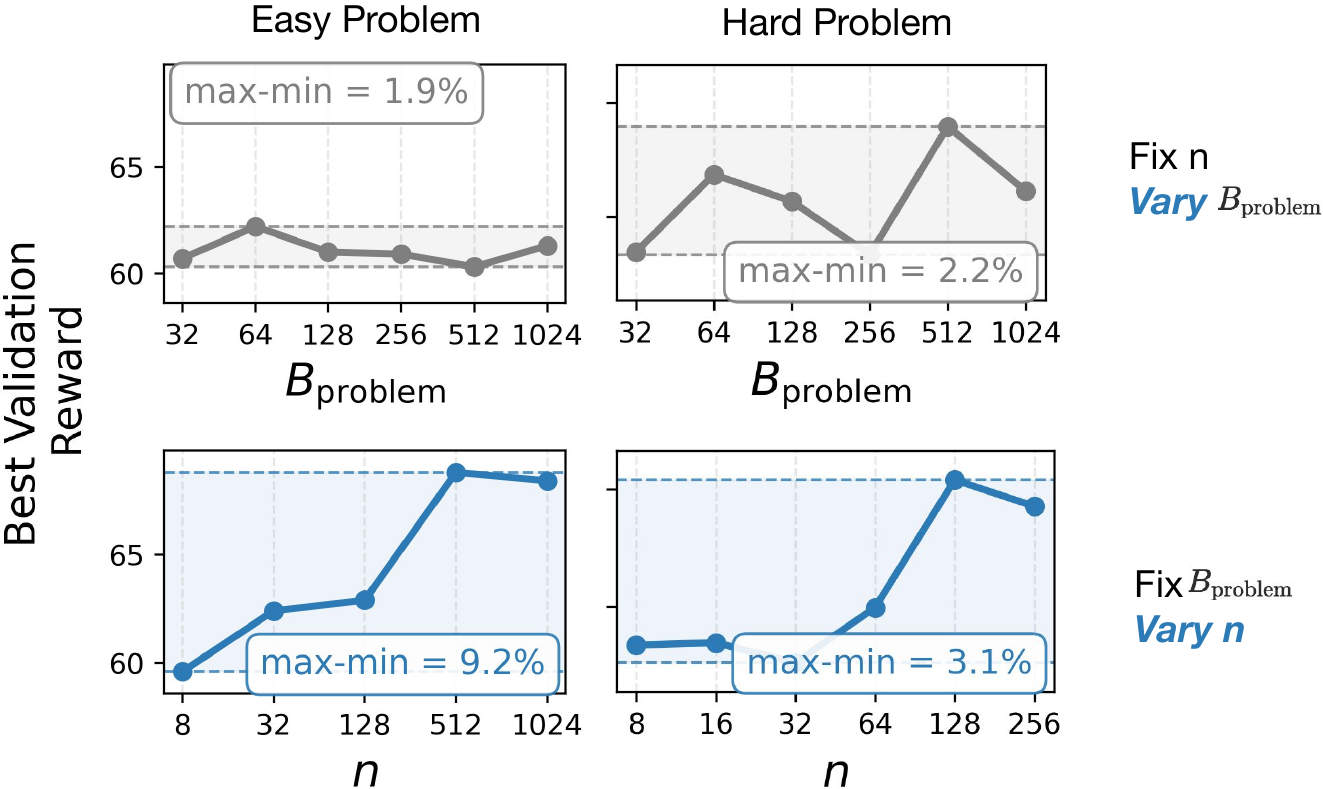}
  \vspace{-0.25cm}
  \caption{\footnotesize{\textbf{Sensitivity of validation reward to $B_\text{p}$ vs. $n$.} Easy \textbf{\textit{(left)}}: The impact of varying $n$ (9.2\% range) shows a clear positive correlation and is significantly larger than varying $B_\text{p}$ (1.9\%). Hard \textbf{\textit{(right)}}: Sensitivity to $B_\text{p}$ (2.2\%) is comparable to $n$ (3.1\%). The fluctuating trend in the top-right plot suggests that $B_\text{p}$ selection introduces optimization instability on hard tasks, explaining the less predictable trends when fixing $B$.}}
  \label{fig:sec3_q2_fixB_varyBprob}
  \vspace{-0.4cm}
\end{wrapfigure}

To make the above argument concrete, we study the effect of varying $B_{\text{p}}$ at fixed $n$, as well as varying $n$ at fixed $B_{\text{p}}$, and assess which hyperparameter more strongly influences performance. On the Easy set, changing $B_{\text{p}}$ has only a marginal effect on validation reward, whereas increasing $n$ leads to substantial gains up to saturation (Figure~\ref{fig:sec3_q2_fixB_varyBprob}, left). This explains the sigmoidal scaling behavior observed earlier: since performance is primarily driven by $n$, increasing $n$ is preferred at larger compute budgets, with $B_{\text{p}}$ decreasing accordingly under a fixed batch size constraint. 
On the Hard set, the picture is more nuanced (Figure~\ref{fig:sec3_q2_fixB_varyBprob}, right). While increasing $n$ remains beneficial, varying $B_{\text{p}}$ produces performance changes of comparable magnitude, and overall sensitivity to both hyperparameters is weaker. As a result, the compute-optimal choice of $n$ is noisier, and at intermediate values of $M$, increasing $B_{\text{p}}$ can yield better performance.

\begin{AIbox}{Key Takeaways}
\begin{enumerate}
    \item With a fixed total batch size $B$, increasing compute favors allocating more rollouts per problem ($n$) and fewer problems per batch ($B_{\text{problem}}$). On the Easy set, this trend follows a clean sigmoidal relationship, since large $B_{\text{problem}}$ overfits due to multi-epoch training.
    \item On the Hard set, the trend is non-monotonic: increasing $B_{\text{problem}}$ may be desirable at intermediate values of $M$. While scaling $n$ remains the more important allocation choice, $B_{\text{problem}}$ must stay above a minimum threshold to avoid incomplete optimization.
\end{enumerate}
\end{AIbox}

\begin{AIbox}{Workflow Prescriptions}
\begin{itemize}
  \item If multi-epoch training on the same problem set is possible, it is preferable to train on fewer problems with a larger per-problem sampling budget $n$. If multi-epoch training is not possible, it may be preferable to include more problems in each batch.

  \item The minimum stable $B_{\text{p}}$ is larger for the Hard set than for the Easy set; on the Easy set, varying $n$ and $B_{\text{p}}$ yields more comparable performance differences.
\end{itemize}
\end{AIbox}

\vspace{-0.1cm}
\subsection{Jointly optimizing $(B_{\text{p}}, n, M)$}
\vspace{-0.1cm}
\label{sec:q3}

Finally, we relax all constraints and jointly optimize the three sampling axes $(B_{\text{p}}, n, M)$ under a fixed total rollout compute budget $C = B_{\text{p}} \cdot n \cdot M$.
The compute-optimal solution is still largely governed by $n$:
\textbf{\emph{the optimal $n^*(C)$ follows a similar sigmoidal scaling with compute (Figure~\ref{fig:q3_nstar})}}.
In contrast, $B_{\text{p}}$ mainly serves as a stability knob and has only a marginal impact on performance within a moderate range.
Practically, we tune $n$ via $n^*(C)$, pick the smallest stable $B_{\text{p}}$, and assign the remaining budget to $M$.
Joint frontiers and sigmoid curves are in Appendix~\ref{app:q3_results}. We also show scaling $n$ improves not only in-domain validation, but also OOD downstream tasks in Appendix~\ref{app:ood_results}~(Figure~\ref{fig:appx_aime24}).

\begin{AIbox}{Key Takeaways}
\begin{enumerate}
  \item When jointly optimizing across all hyperparameters $(n, B_{\text{problem}}, M)$, the compute-optimal value of $n$ still increases with $C$, consistent with the findings from Questions~1 and~2.

  \item The best total rollout size $B$ generally increases as $C$ increases, although the compute-optimal $B_{\text{p}}$ can often be chosen to be budget-agnostic.

\end{enumerate}
\end{AIbox}

\vspace{-0.2cm}
\section{Role of Base Model and Prompt Set}
\label{sec:bigger_picture}
\vspace{-0.1cm}

\begin{wrapfigure}{r}{0.55\textwidth}
  \centering
  \vspace{-0.7cm}
  \includegraphics[width=\linewidth]{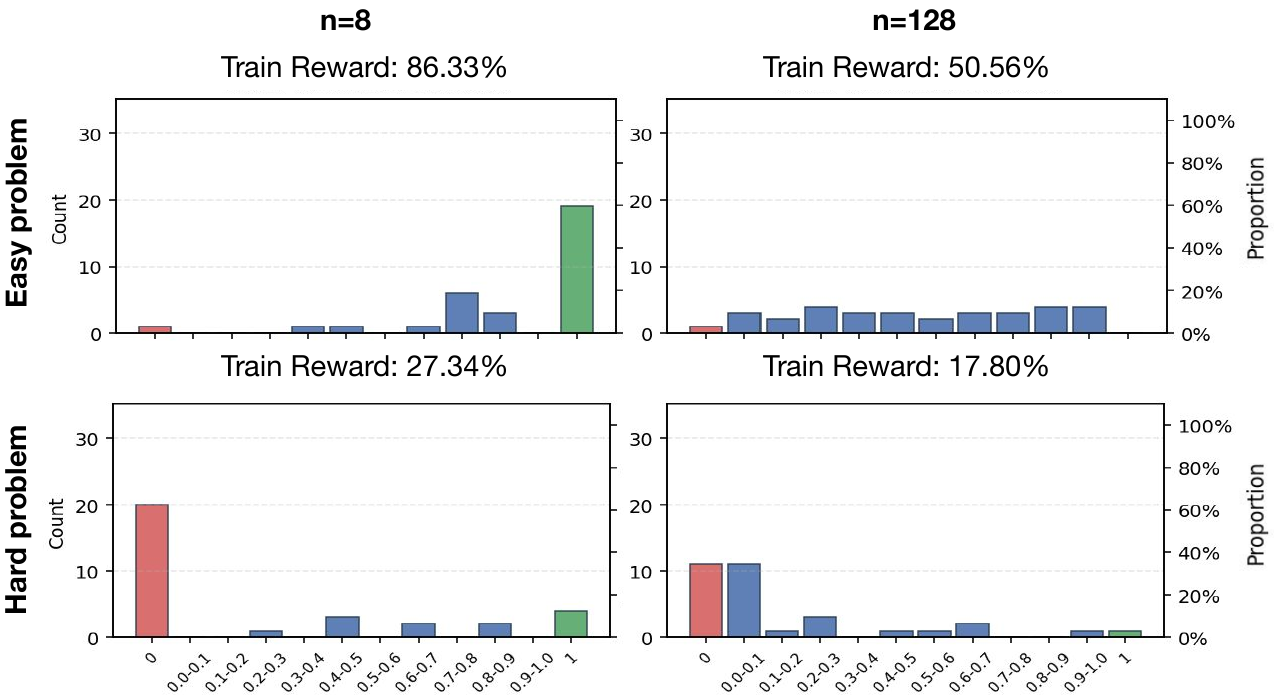}
  \vspace{-0.6cm}
  \caption{\footnotesize{\textbf{Training reward distributions on Easy and Hard sets at a matched compute level ($n=8$ vs. $n=128$).}
  (1) Interference exists: On the Easy set (initial pass rate 0.3-0.6), optimization sacrifices some problems, leaving a non-zero fraction unsolved after training.
  (2) Easy set: Larger $n$ results in a more uniform distribution of pass rates, avoiding polarized outcomes seen in smaller $n$.
  (3) Hard set: Larger $n$ improves coverage (reducing zero fraction), while smaller $n$ sharpens performance on a subset.}}
  \label{fig:train_pass1_dist}
  \vspace{-0.6cm}
\end{wrapfigure}

Having seen that the compute-optimal number of rollouts $n$ increases with sampling compute $C$ on both Easy and Hard sets, it is natural to ask whether this behavior extends to other prompt distributions and base models. We also note that this qualitative trend is not specific to the GRPO algorithm considered here, and appears under other algorithmic variants~(PPO~\cite{schulman2017proximal} and CISPO~\cite{minimax2025minimaxm1scalingtesttimecompute}) as well in Appendix~\ref{app:algos} Figure~\ref{fig:appx_other_algos}.

\vspace{-0.2cm}
\subsection{Scaling $n$ Addresses Interference}
\vspace{-0.1cm}
If we were given a multi-armed bandit problem, in a tabular setting, the compute-optimal scaling strategy would prescribe increasing $M$ (sequential updates) over using a higher $n$ (as discussed in Appendix~\ref{app:base_case_one_problem}). However, this theoretical prediction contradicts our empirical findings that show scaling $n$ is better. In this section, we argue that this gap arises due to \emph{\textbf{interference}} across problems~\cite{arxiv-org-1904-11455,qu2026popelearningreasonhard}.

\begin{figure}[t]
  \centering
  \includegraphics[width=0.85\linewidth]{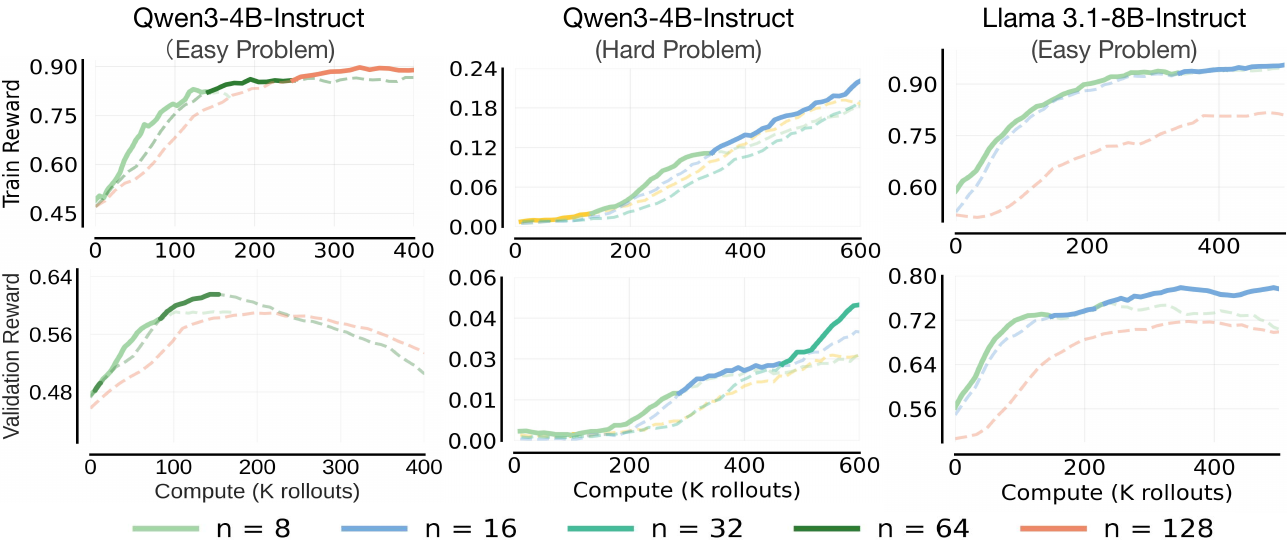}
  \caption{\footnotesize{\textbf{Generalizing $n$ scaling trends to other models.} We observe increasing $n$ boosts returns at high compute across all settings, while optimal $n$ saturates differently.}}
  \label{fig:base_model_generalization}
  \vspace{-0.3cm}
\end{figure}

When multiple problems are trained jointly, gradient updates can interfere, possibly causing uneven learning across problems and degradation on previously solvable problems. In this regime, a larger $n$ is preferable to increasing $M$, since more rollouts yield more uniform updates across problems per step and improve learning efficiency. This shifts the compute-optimal balance toward parallel sampling rather than sequential optimization, mitigating interference and improving learning efficiency.

\textbf{Evaluating interference.}
To quantify interference, we analyze the training-set pass@1 distribution across problems under matched compute budgets ($n \cdot M$). Even on the Easy set, a non-trivial fraction of problems end training with pass@1 close to zero, indicating uneven progress across problems. Under the same compute budget, larger values of $n$ yield a less skewed distribution and more uniform improvements (Figure~\ref{fig:train_pass1_dist}). 
A similar pattern appears on the Hard set: smaller $n$ optimizes on a subset of problems while leaving many unsolved, whereas larger $n$ reduces the zero-pass fraction. Overall, increasing $n$ mitigates interference by distributing updates more evenly across problems, explaining why it is preferred.

\textbf{Compute-optimal $n$ scaling generalizes for different base models.} As shown in Figure~\ref{fig:base_model_generalization}, larger $n$ values consistently outperform the baseline ($n=8$) at high compute budgets for both Qwen3-4B-Instruct and Llama 3.1-8B-Instruct on their Easy and Hard sets. These results are consistent with our main compute-optimal findings. However, the optimal values of $n$ vary across model--dataset pairs. One plausible explanation is that different base models begin with different effective competence on the target problem distribution, which changes the available reward density and the range of compute over which larger $n$ remains beneficial.
We also observe that, on easy problems, validation reward for both models saturates or degrades at $n=128$, even while the \emph{\textbf{training reward continues to rise}}. We attribute this divergence to the train--test gap (overfitting), discussed next.


\vspace{-0.2cm}
\subsection{Train-Test Gap}
\label{sec:train_test}
\vspace{-0.1cm}

\begin{figure}[!tb]
  \centering
  \includegraphics[width=0.65\linewidth]{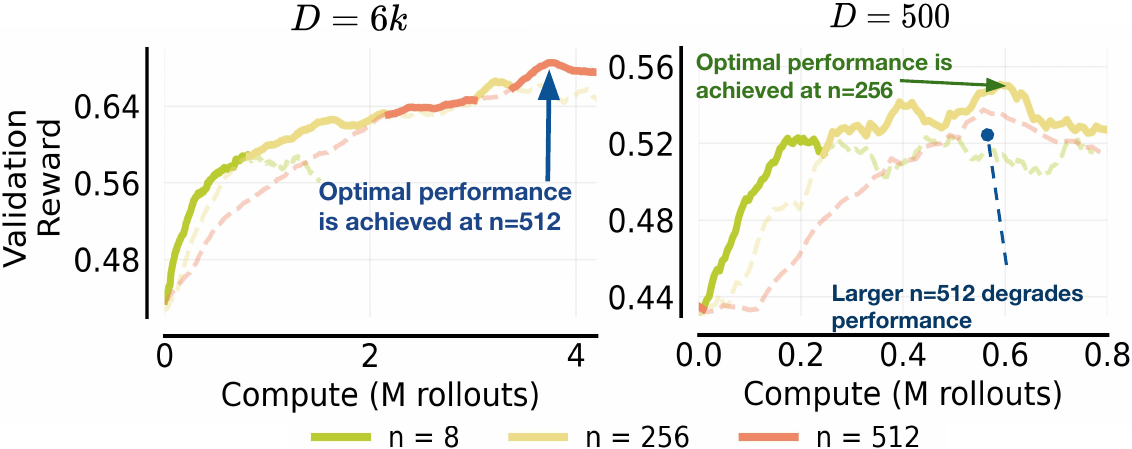}
  \caption{\footnotesize{\textbf{Impact of data size ($D$).} With more data ($D=6\text{k}$; \textbf{\textit{left}}), performance scales up to $n=512$. With small data ($D=500$; \textbf{\textit{right}}), the frontier saturates at smaller $n=256$, and scaling further to $n=512$ leads to overfitting and degradation.}}
  \label{fig:varyD_overfitting}
\end{figure}

Our scaling results use validation metrics, even though optimization dynamics are driven by the training set. Thus, scaling laws on the validation set require sustained transfer from training to test. When the prompt set is too small, training may overfit early, so larger $n$ may no longer appear compute-optimal even at high budgets, as additional training fails to improve validation performance. Figure~\ref{fig:varyD_overfitting} shows that when we vary the prompt set size $D$, the compute-optimal $n$ caps at smaller values for smaller $D$. This is expected: validation reward degrades under prolonged training due to overfitting, preventing larger $n$ from appearing on the frontier. As a result, the compute-optimal allocation for training performance may differ from that for validation, especially at large compute budgets.

\begin{figure}[t]
  \centering
  \vspace{-0.2cm}
  \includegraphics[width=0.8\linewidth]{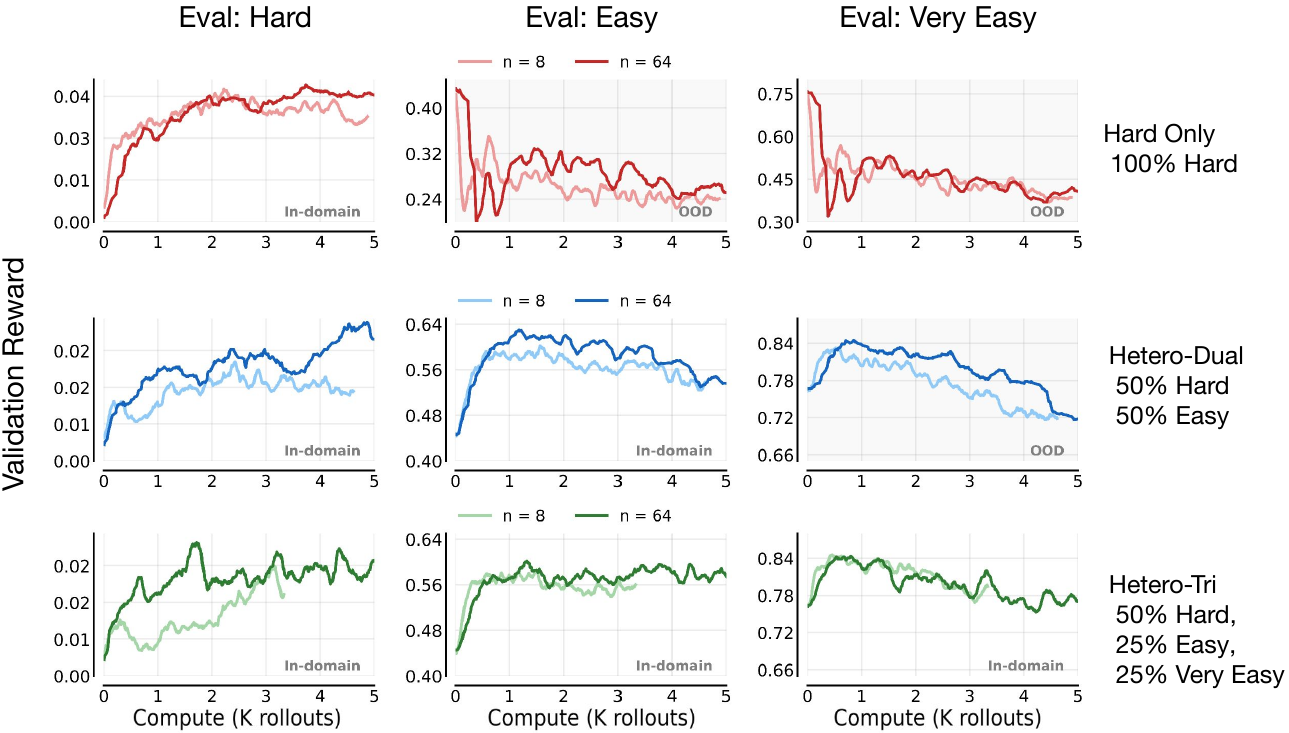}
\caption{\footnotesize{\textbf{Results across difficulty levels for small ($n=8$) and large ($n=64$) rollout budgets under different training data distributions} (5K total samples) using Qwen2.5-7B-Instruct. We consider Hard ($\text{pass}@128=0$), Easy ($\text{pass}@128 \in [0.3,0.6]$), and Very Easy ($\text{pass}@128 \in [0.6,0.9]$) problems. Rows correspond to Hard Only, Heterogeneous-Dual Mix (50\% Hard, 50\% Easy), and Heterogeneous-Tri Mix (50\% Hard, 25\% Easy, 25\% Very Easy; the J-shaped distribution from Polaris). \emph{Across distributions, larger $n$ consistently performs better at higher compute in in-domain evaluations}, except on the Very Easy evaluation set, where the task is likely too easy for additional compute to matter. \emph{Training only on Hard data causes substantial catastrophic forgetting on Easy and Very Easy problems}, while mixing Easy data largely mitigates this effect with only a small drop on Hard performance. In contrast, adding Very Easy data does not help and can hurt both Easy and Hard performance.}}
  \label{fig:skewed_dist}
  \vspace{-0.25cm}
\end{figure}

\vspace{-0.2cm}
\subsection{Other Data Compositions}
\label{sec:dataset_skew}
\vspace{-0.15cm}

Finally, we train on heterogeneous mixtures of Easy and Hard problems (Figure~\ref{fig:skewed_dist}), as well as an ``extra hard'' set where the base model attains $\text{pass}@128=0$. These mixtures induce different skewness and thereby alter the rate at which $\text{pass}@1$ improves in training. Despite this variation, we observe a {consistent crossover} trend that larger $n$ outperforms smaller $n$ on validation sets. 
The compute ranges where small $n$ is optimal are different.
This suggests the rate of $\text{pass}@1$ improvement controls both the compute range over which a given $n$ is optimal and the minimum compute-optimal $n$. 
Crucially, we note that our central finding remains unchanged: \textbf{\emph{larger compute budgets $C$ support larger compute-optimal values of $n$, even on skewed training mixtures}}.

\begin{AIbox}{Key Takeaways}
\begin{enumerate}
  \item \emph{Interference favors more parallel rollouts.} Unlike the tabular case, interference across problems makes larger $n$ beneficial, and both $n$ and $M$ increase with compute.

  \item \emph{Small training sets induce early saturation.} Running Multiple epochs of training on a given problem set can result in overfitting, causing early saturation on validation metrics and shifting the optimum toward larger $B_{\text{p}}$ and smaller $n, M$.

  \item \emph{Scaling rules are broadly transferable.} Similar compute-allocation trends hold across base models and data compositions, though the optimal range of $n$ still depends on the dataset.
\end{enumerate}
\end{AIbox}

\vspace{-0.2cm}
\section{Related Work}
\vspace{-0.15cm}

Scaling laws are well established for pretraining~\citep{arxiv-org-1712-00409,arxiv-org-2001-08361,arxiv-org-2203-15556}, but predicting RL behaviors is more challenging due to coupled data collection and optimization. Prior work reports approximate power-law scaling in controlled RL settings such as board games and single-agent deep RL~\citep{arxiv-org-2104-03113,arxiv-org-2301-13442}, and characterizes compute-data trade-offs and Pareto frontiers in value-based RL~\citep{value-scaling-github-io-value-scaling-github-io,rybkin2025valuebaseddeeprlscales}. 

Whether such predictability extends to LLM RL remains unclear, as experience is generated on-policy at high cost and scaling behavior depends on recipe-level stability. Recent studies make progress by extending on-policy RL under fixed pipelines and observing sigmoidal reward--compute curves~\citep{arxiv-org-2510-13786}, or varying model size~\citep{arxiv-org-2509-25300}. However, instabilities such as entropy collapse or policy drift often require stabilizers including KL, clipping, or resets~\citep{arxiv-org-2505-22617,arxiv-org-2510-01180}. 

There have also been works exploring scaling LLM RL along separate axes of compute. On the axis of sequential scaling, DeepSeek-R1~\citep{guo2025deepseek} showed that RLVR could largely improve reasoning capability, while ProRL~\citep{liu2025prorl} explicitly highlighted the importance of prolonged RL training; similarly, works such as DAPO~\citep{yu2025dapo} and OpenReasonerZero~\citep{hu2025open}, though not framed as scaling-law studies, naturally scale along sequential updates until reward convergence. On the axis of parallel rollouts per sample, BroRL~\citep{arxiv-org-2510-01180} studied rollout width and showed that broader exploration can overcome plateaus arising from purely sequential scaling, while KnapSackRL~\citep{li2025knapsack} considered adaptive budget allocation instead of uniform sampling. While the impact of batch size has been studied in pretraining contexts~\citep{mccandlish2018empirical,gray2023efficient,zhangdoes}, there is still limited work systematically scaling problem batch size in the LLM RL setting. Other dimensions of scaling LLM RL include scaling problem sets~\citep{arxiv-org-2506-14965}, environments~\citep{zeng2025rlve}, and model size~\citep{arxiv-org-2509-25300}. 

As a result, existing work largely \emph{describes} scaling along fixed recipes or studies individual axes, whereas practitioners face a \emph{budget allocation} problem: how to allocate a fixed sampling budget across various hyperparameters in LLM RL. We therefore study RL scaling laws as \emph{prescriptive allocation rules}, using compute-optimal analysis over $(B_{\text{p}}, n, M)$ under stable recipes.

\section{Discussion and Conclusion}
\label{sec:discussion_conclusion}
\vspace{-0.1cm}

A central takeaway from this work is that healthy RL recipes are inherently dependent on the prompt distribution, and that RL training behavior emerges from the interaction between the base model, the prompt set, and the available compute budget. This dependence manifests directly in how optimal hyperparameters scale with compute, so that the same algorithm can exhibit qualitatively different scaling behavior on easy versus hard problem sets. On easier problems, increasing parallel rollout compute primarily improves sharpening and robustness, whereas on harder problems the dominant effect is expanded coverage through improved discovery of rare successful trajectories. While trends in compute-optimal hyperparameters are often consistent when measured using average reward, they can diverge substantially under alternative metrics such as best@k and worst@k. This sensitivity to both data difficulty and evaluation metric highlights an important difference from supervised learning, where scaling behavior is often more uniform once model size is fixed. 

Framing RL training as a compute-constrained allocation problem makes this dependence operational: across the settings we study, the compute-optimal number of parallel rollouts per problem ($n$) increases with the available sampling budget and eventually saturates, while the number of problems per batch ($B_{\text{p}}$) primarily acts as a stability knob with weaker effects once it lies in a moderate range. Under fixed batch-size constraints, this yields a practical rule: favor larger $B_{\text{p}}$ when only a small number of sequential updates is possible, and shift compute toward larger $n$ as the available budget grows. Joint optimization over $(B_{\text{p}}, n, M)$ leads to a similar conclusion: the allocation frontier is governed primarily by $n$, with the remaining budget best assigned to stable choices of $B_{\text{p}}$ and then to $M$.

\textbf{Directions for future work.} Our analysis also surfaces an important open challenge: interference across problems. In an idealized single-problem setting, one might expect clean exponential improvements with increasing sampling compute. In practice, however, RL is performed over mixtures of problems, where progress on some tasks can interfere with learning on others. This population-level interference alters both the coefficients and the effective hyperparameter values in observed scaling laws.

Another promising direction for future work is to identify sufficient statistics early on in a training run that capture the degree of interference across problems, enabling more accurate predictions of how additional compute will translate into subsequent learning progress. Tracking changes in the pass@1 distribution through training provides a natural starting point for studying such interference. More broadly, developing predictive models based on a small set of statistics summarizing the pass@1 landscape may enable approximate closed-form rules for compute-optimal hyperparameters that generalize across base models and prompt distributions.

\vspace{-0.2cm}
\section*{Acknowledgements}
\vspace{-0.2cm}
We thank Oleg Rybkin, Apurva Gandhi, Charlie Snell, Matthew Yang, Rishabh Agarwal, Sang Michael Xie, Junlong Li, Zora Wang, and other members of the CMU AIRe lab for their thoughtful feedback and discussions. We also thank Chengyu Dong, Mikhail Yurochkin, Rupesh Srivastava, Joel Hestness, and Gavia Gray for early discussions on RL scaling in LLM. We also gratefully acknowledge the Orchard cluster at the FLAME center of CMU for providing computational resources that supported a part of this work.

\newpage
\bibliography{main}

@misc{yang2022tensorprogramsvtuning,
      title={Tensor Programs V: Tuning Large Neural Networks via Zero-Shot Hyperparameter Transfer}, 
      author={Greg Yang and Edward J. Hu and Igor Babuschkin and Szymon Sidor and Xiaodong Liu and David Farhi and Nick Ryder and Jakub Pachocki and Weizhu Chen and Jianfeng Gao},
      year={2022},
      eprint={2203.03466},
      archivePrefix={arXiv},
      primaryClass={cs.LG},
      url={https://arxiv.org/abs/2203.03466}, 
}

@article{arxiv-org-1712-00409,
  title={Deep learning scaling is predictable, empirically},
  author={Hestness, Joel and Narang, Sharan and Ardalani, Newsha and Diamos, Gregory and Jun, Heewoo and Kianinejad, Hassan and Patwary, Md Mostofa Ali and Yang, Yang and Zhou, Yanqi},
  journal={arXiv preprint arXiv:1712.00409},
  year={2017}
}

@article{arxiv-org-2001-08361,
  title={Scaling laws for neural language models},
  author={Kaplan, Jared and McCandlish, Sam and Henighan, Tom and Brown, Tom B and Chess, Benjamin and Child, Rewon and Gray, Scott and Radford, Alec and Wu, Jeffrey and Amodei, Dario},
  journal={arXiv preprint arXiv:2001.08361},
  year={2020}
}

@misc{rybkin2025valuebaseddeeprlscales,
      title={Value-Based Deep RL Scales Predictably}, 
      author={Oleh Rybkin and Michal Nauman and Preston Fu and Charlie Snell and Pieter Abbeel and Sergey Levine and Aviral Kumar},
      year={2025},
      eprint={2502.04327},
      archivePrefix={arXiv},
      primaryClass={cs.LG},
      url={https://arxiv.org/abs/2502.04327}, 
}

@misc{qu2026popelearningreasonhard,
      title={POPE: Learning to Reason on Hard Problems via Privileged On-Policy Exploration}, 
      author={Yuxiao Qu and Amrith Setlur and Virginia Smith and Ruslan Salakhutdinov and Aviral Kumar},
      year={2026},
      eprint={2601.18779},
      archivePrefix={arXiv},
      primaryClass={cs.LG},
      url={https://arxiv.org/abs/2601.18779}, 
}

@misc{snell2024scalingllmtesttimecompute,
      title={Scaling LLM Test-Time Compute Optimally can be More Effective than Scaling Model Parameters}, 
      author={Charlie Snell and Jaehoon Lee and Kelvin Xu and Aviral Kumar},
      year={2024},
      eprint={2408.03314},
      archivePrefix={arXiv},
      primaryClass={cs.LG},
      url={https://arxiv.org/abs/2408.03314}, 
}

@article{arxiv-org-2203-15556,
  title={Training compute-optimal large language models},
  author={Hoffmann, Jordan and Borgeaud, Sebastian and Mensch, Arthur and Buchatskaya, Elena and Cai, Trevor and Rutherford, Eliza and Casas, Diego de Las and Hendricks, Lisa Anne and Welbl, Johannes and Clark, Aidan and others},
  journal={arXiv preprint arXiv:2203.15556},
  year={2022}
}

@article{arxiv-org-2104-03113,
  title={Scaling scaling laws with board games},
  author={Jones, Andy L},
  journal={arXiv preprint arXiv:2104.03113},
  year={2021}
}

@article{arxiv-org-2301-13442,
  title={Scaling laws for single-agent reinforcement learning},
  author={Hilton, Jacob and Tang, Jie and Schulman, John},
  journal={arXiv preprint arXiv:2301.13442},
  year={2023}
}

@article{arxiv-org-2510-13786,
  title={The art of scaling reinforcement learning compute for llms},
  author={Khatri, Devvrit and Madaan, Lovish and Tiwari, Rishabh and Bansal, Rachit and Duvvuri, Sai Surya and Zaheer, Manzil and Dhillon, Inderjit S and Brandfonbrener, David and Agarwal, Rishabh},
  journal={arXiv preprint arXiv:2510.13786},
  year={2025}
}

@article{arxiv-org-2509-25300,
  title={Scaling Behaviors of LLM Reinforcement Learning Post-Training: An Empirical Study in Mathematical Reasoning},
  author={Tan, Zelin and Geng, Hejia and Yu, Xiaohang and Zhang, Mulei and Wan, Guancheng and Zhou, Yifan and He, Qiang and Xue, Xiangyuan and Zhou, Heng and Fan, Yutao and others},
  journal={arXiv preprint arXiv:2509.25300},
  year={2025}
}

@article{arxiv-org-2402-03300-2,
  title={Deepseekmath: Pushing the limits of mathematical reasoning in open language models},
  author={Shao, Zhihong and Wang, Peiyi and Zhu, Qihao and Xu, Runxin and Song, Junxiao and Bi, Xiao and Zhang, Haowei and Zhang, Mingchuan and Li, YK and Wu, Yang and others},
  journal={arXiv preprint arXiv:2402.03300},
  year={2024}
}

@article{arxiv-org-2505-22617,
  title={The entropy mechanism of reinforcement learning for reasoning language models},
  author={Cui, Ganqu and Zhang, Yuchen and Chen, Jiacheng and Yuan, Lifan and Wang, Zhi and Zuo, Yuxin and Li, Haozhan and Fan, Yuchen and Chen, Huayu and Chen, Weize and others},
  journal={arXiv preprint arXiv:2505.22617},
  year={2025}
}

@article{arxiv-org-2506-14965,
  title={Revisiting Reinforcement Learning for LLM Reasoning from A Cross-Domain Perspective},
  author={Cheng, Zhoujun and Hao, Shibo and Liu, Tianyang and Zhou, Fan and Xie, Yutao and Yao, Feng and Bian, Yuexin and Zhuang, Yonghao and Dey, Nilabjo and Zha, Yuheng and others},
  journal={arXiv preprint arXiv:2506.14965},
  year={2025}
}

@article{arxiv-org-2510-01180,
  title={Brorl: Scaling reinforcement learning via broadened exploration},
  author={Hu, Jian and Liu, Mingjie and Lu, Ximing and Wu, Fang and Harchaoui, Zaid and Diao, Shizhe and Choi, Yejin and Molchanov, Pavlo and Yang, Jun and Kautz, Jan and others},
  journal={arXiv preprint arXiv:2510.01180},
  year={2025}
}

@inproceedings{mei2023stochastic,
  title={Stochastic gradient succeeds for bandits},
  author={Mei, Jincheng and Zhong, Zixin and Dai, Bo and Agarwal, Alekh and Szepesvari, Csaba and Schuurmans, Dale},
  booktitle={International Conference on Machine Learning},
  pages={24325--24360},
  year={2023},
  organization={PMLR}
}

@article{arxiv-org-1904-11455,
  title={Ray interference: a source of plateaus in deep reinforcement learning},
  author={Schaul, Tom and Borsa, Diana and Modayil, Joseph and Pascanu, Razvan},
  journal={arXiv preprint arXiv:1904.11455},
  year={2019}
}

@article{arxiv-org-2502-17578,
  title={How Do Large Language Monkeys Get Their Power (Laws)?},
  author={Schaeffer, Rylan and Kazdan, Joshua and Hughes, John and Juravsky, Jordan and Price, Sara and Lynch, Aengus and Jones, Erik and Kirk, Robert and Mirhoseini, Azalia and Koyejo, Sanmi},
  journal={arXiv preprint arXiv:2502.17578},
  year={2025}
}

@article{value-scaling-github-io-value-scaling-github-io,
  title = {Scaling Laws for Value-Based RL},
  author = {Fu, Preston and Rybkin, Oleh and Kumar, Aviral},
  journal = {value-scaling.github.io},
  year = {2025},
  month = {September},
  url = "https://value-scaling.github.io/"
}

@misc{arxiv-org-2406-14532,
      title={RL on Incorrect Synthetic Data Scales the Efficiency of LLM Math Reasoning by Eight-Fold}, 
      author={Amrith Setlur and Saurabh Garg and Xinyang Geng and Naman Garg and Virginia Smith and Aviral Kumar},
      year={2024},
      eprint={2406.14532},
      archivePrefix={arXiv},
      primaryClass={cs.LG},
      url={https://arxiv.org/abs/2406.14532}, 
}

@misc{yao2025offpolicy,
  title = {Your Efficient RL Framework Secretly Brings You Off-Policy RL Training},
  url = {https://fengyao.notion.site/off-policy-rl},
  author = {Yao, Feng and Liu, Liyuan and Zhang, Dinghuai and Dong, Chengyu and Shang, Jingbo and Gao, Jianfeng},
  journal = {Feng Yao's Notion},
  year = {2025},
  month = aug,
}

@inproceedings{loshchilovdecoupled,
  title={Decoupled Weight Decay Regularization},
  author={Loshchilov, Ilya and Hutter, Frank},
  booktitle={International Conference on Learning Representations}
}

@article{liu2025prorl,
  title={Prorl: Prolonged reinforcement learning expands reasoning boundaries in large language models},
  author={Liu, Mingjie and Diao, Shizhe and Lu, Ximing and Hu, Jian and Dong, Xin and Choi, Yejin and Kautz, Jan and Dong, Yi},
  journal={arXiv preprint arXiv:2505.24864},
  year={2025}
}

@article{yu2025dapo,
  title={Dapo: An open-source llm reinforcement learning system at scale},
  author={Yu, Qiying and Zhang, Zheng and Zhu, Ruofei and Yuan, Yufeng and Zuo, Xiaochen and Yue, Yu and Dai, Weinan and Fan, Tiantian and Liu, Gaohong and Liu, Lingjun and others},
  journal={arXiv preprint arXiv:2503.14476},
  year={2025}
}

@article{mccandlish2018empirical,
  title={An empirical model of large-batch training},
  author={McCandlish, Sam and Kaplan, Jared and Amodei, Dario and Team, OpenAI Dota},
  journal={arXiv preprint arXiv:1812.06162},
  year={2018}
}

@article{li2025knapsack,
  title={Knapsack rl: Unlocking exploration of llms via optimizing budget allocation},
  author={Li, Ziniu and Chen, Congliang and Yang, Tianyun and Ding, Tian and Sun, Ruoyu and Zhang, Ge and Huang, Wenhao and Luo, Zhi-Quan},
  journal={arXiv preprint arXiv:2509.25849},
  year={2025}
}

@article{guo2025deepseek,
  title={Deepseek-r1: Incentivizing reasoning capability in llms via reinforcement learning},
  author={Guo, Daya and Yang, Dejian and Zhang, Haowei and Song, Junxiao and Zhang, Ruoyu and Xu, Runxin and Zhu, Qihao and Ma, Shirong and Wang, Peiyi and Bi, Xiao and others},
  journal={arXiv preprint arXiv:2501.12948},
  year={2025}
}

@inproceedings{gray2023efficient,
  title={Efficient and approximate per-example gradient norms for gradient noise scale},
  author={Gray, Gavia and Samar, Anshul and Hestness, Joel},
  booktitle={Workshop on Advancing Neural Network Training: Computational Efficiency, Scalability, and Resource Optimization (WANT@ NeurIPS 2023)},
  year={2023}
}

@inproceedings{zhangdoes,
  title={How Does Critical Batch Size Scale in Pre-training?},
  author={Zhang, Hanlin and Morwani, Depen and Vyas, Nikhil and Wu, Jingfeng and Zou, Difan and Ghai, Udaya and Foster, Dean and Kakade, Sham M},
  booktitle={The Thirteenth International Conference on Learning Representations}
}

@article{zeng2025rlve,
  title={Rlve: Scaling up reinforcement learning for language models with adaptive verifiable environments},
  author={Zeng, Zhiyuan and Ivison, Hamish and Wang, Yiping and Yuan, Lifan and Li, Shuyue Stella and Ye, Zhuorui and Li, Siting and He, Jacqueline and Zhou, Runlong and Chen, Tong and others},
  journal={arXiv preprint arXiv:2511.07317},
  year={2025}
}

@article{hu2025open,
  title={Open-reasoner-zero: An open source approach to scaling up reinforcement learning on the base model},
  author={Hu, Jingcheng and Zhang, Yinmin and Han, Qi and Jiang, Daxin and Zhang, Xiangyu and Shum, Heung-Yeung},
  journal={arXiv preprint arXiv:2503.24290},
  year={2025}
}

@article{sheng2024hybridflow,
  title={HybridFlow: A Flexible and Efficient RLHF Framework},
  author={Sheng, Guangming and Zhang, Chi and Ye, Zilingfeng and Wu, Xibin and Zhang, Wang and Zhang, Ru and Peng, Yanghua and Lin, Haibin and Wu, Chuan},
  journal={arXiv preprint arXiv:2409.19256},
  year={2024}
}

@misc{minimax2025minimaxm1scalingtesttimecompute,
      title={MiniMax-M1: Scaling Test-Time Compute Efficiently with Lightning Attention}, 
      author={MiniMax and : and Aili Chen and Aonian Li and Bangwei Gong and Binyang Jiang and Bo Fei and Bo Yang and Boji Shan and Changqing Yu and Chao Wang and Cheng Zhu and Chengjun Xiao and Chengyu Du and Chi Zhang and Chu Qiao and Chunhao Zhang and Chunhui Du and Congchao Guo and Da Chen and Deming Ding and Dianjun Sun and Dong Li and Enwei Jiao and Haigang Zhou and Haimo Zhang and Han Ding and Haohai Sun and Haoyu Feng and Huaiguang Cai and Haichao Zhu and Jian Sun and Jiaqi Zhuang and Jiaren Cai and Jiayuan Song and Jin Zhu and Jingyang Li and Jinhao Tian and Jinli Liu and Junhao Xu and Junjie Yan and Junteng Liu and Junxian He and Kaiyi Feng and Ke Yang and Kecheng Xiao and Le Han and Leyang Wang and Lianfei Yu and Liheng Feng and Lin Li and Lin Zheng and Linge Du and Lingyu Yang and Lunbin Zeng and Minghui Yu and Mingliang Tao and Mingyuan Chi and Mozhi Zhang and Mujie Lin and Nan Hu and Nongyu Di and Peng Gao and Pengfei Li and Pengyu Zhao and Qibing Ren and Qidi Xu and Qile Li and Qin Wang and Rong Tian and Ruitao Leng and Shaoxiang Chen and Shaoyu Chen and Shengmin Shi and Shitong Weng and Shuchang Guan and Shuqi Yu and Sichen Li and Songquan Zhu and Tengfei Li and Tianchi Cai and Tianrun Liang and Weiyu Cheng and Weize Kong and Wenkai Li and Xiancai Chen and Xiangjun Song and Xiao Luo and Xiao Su and Xiaobo Li and Xiaodong Han and Xinzhu Hou and Xuan Lu and Xun Zou and Xuyang Shen and Yan Gong and Yan Ma and Yang Wang and Yiqi Shi and Yiran Zhong and Yonghong Duan and Yongxiang Fu and Yongyi Hu and Yu Gao and Yuanxiang Fan and Yufeng Yang and Yuhao Li and Yulin Hu and Yunan Huang and Yunji Li and Yunzhi Xu and Yuxin Mao and Yuxuan Shi and Yuze Wenren and Zehan Li and Zelin Li and Zhanxu Tian and Zhengmao Zhu and Zhenhua Fan and Zhenzhen Wu and Zhichao Xu and Zhihang Yu and Zhiheng Lyu and Zhuo Jiang and Zibo Gao and Zijia Wu and Zijian Song and Zijun Sun},
      year={2025},
      eprint={2506.13585},
      archivePrefix={arXiv},
      primaryClass={cs.CL},
      url={https://arxiv.org/abs/2506.13585}, 
}

@article{schulman2017proximal,
  title={Proximal policy optimization algorithms},
  author={Schulman, John and Wolski, Filip and Dhariwal, Prafulla and Radford, Alec and Klimov, Oleg},
  journal={arXiv preprint arXiv:1707.06347},
  year={2017}
}

\newpage
\appendix
\onecolumn
\part*{Appendices}

\section{Detailed Experiment Setup}
\label{app:exp_detail}
 \textbf{Recipe ablation setup.} We use Qwen2.5-7B-Instruct (max length 8,192) with GRPO. 
For regularizer ablations, we fix $B_\text{p}=256$ and $n=16$. On both Easy and Hard sets, we ablate KL and entropy regularization and the zero-variance filter (including applying it selectively to loss terms). 
For LR scaling, we use AdamW~\citep{loshchilovdecoupled} with a 10-step linear warmup followed by a constant schedule. We establish a base LR anchor at $B_\text{p}=128, n=8$ ($B=1,024$) via grid search. We then scale to $n=64$ ($B=8,192$) to compare three scaling rules: (1) constant, (2) linear, and (3) square-root scaling.

\textbf{Zero-variance filtering} is employed in recent works~\citep{arxiv-org-2510-13786} to exclude prompts with identical rollout rewards from loss in GRPO. This mechanism increases effective batch size and prevents applying regularizers to zero-gradient trajectories, a crucial feature for hard problems where exploration naturally drives high entropy. However, our experiments (Figure~\ref{fig:sec2_kl_ent_ablation}) show that even when filtering is \textit{applied to KL and entropy terms}, instability and entropy explosion persist, though mitigated, when rare positives are sampled. Since removing regularization entirely yields the most stable dynamics, we employ KL+entropy regularization only on the Easy set and omit them on the Hard set to avoid instability.

\textbf{Main experiment setup.} We train Qwen2.5-7B-Instruct with on-policy updates using the optimized recipe above. The learning rate scales proportionally to $\sqrt{B}$ (base 1e-6 at $B=1024$). Based on ablation results, KL and entropy regularization are enabled for the Easy set but disabled for the Hard set. We fix temperature to 0.6 and top-$p$ to 1.0. We use the GRPO algorithm and Truncated Importance Sampling (TIS~\citep{yao2025offpolicy}) to mitigate training-inference logit mismatch. We use the veRL~\citep{sheng2024hybridflow} framework to conduct all RL experiments.

\begin{figure}[!ht]
    \centering
    \includegraphics[width=0.9\linewidth]{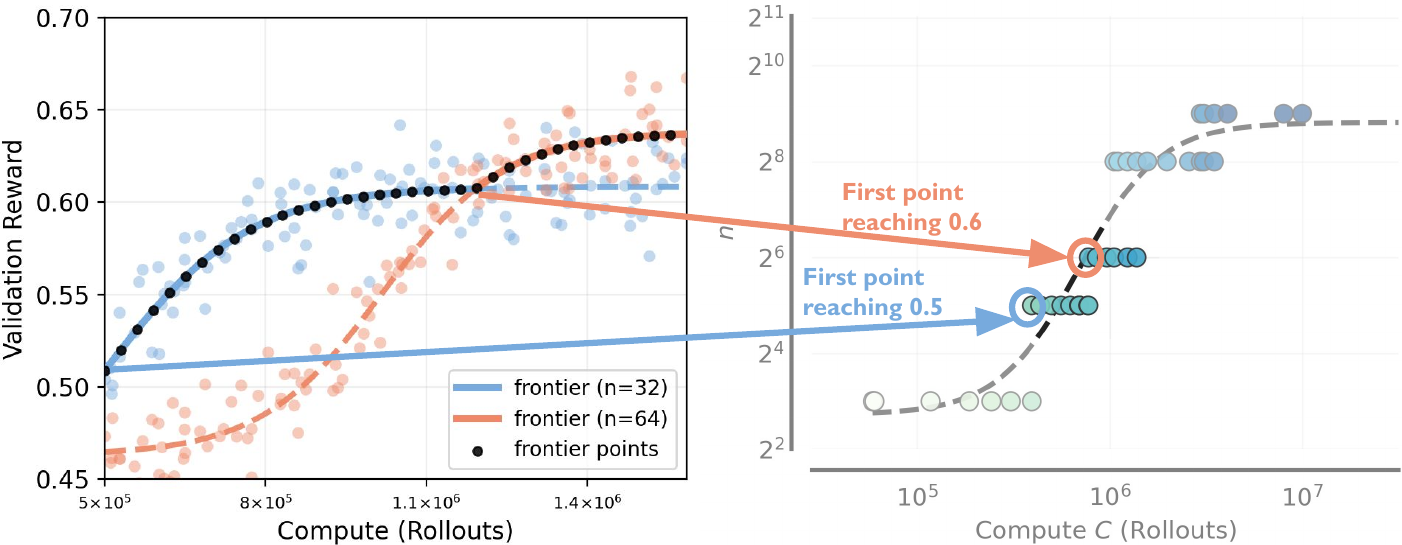}
    \caption{\footnotesize{\textbf{Demonstrations of frontier point detection for each $n$}.
    \textbf{(Left)} Validation reward trajectories plotted against compute (rollouts) for varying population sizes ($n=32$ in blue, $n=64$ in red). Scatter points show raw data; dashed curves show smoothed trends. Arrows illustrate the ``record-breaking'' extraction process, identifying the earliest compute step where reward crosses a discretized threshold (e.g., $0.5$ or $0.6$). In practice, we employ finer reward bins (e.g., $0.005$) tailored to task difficulty.
    \textbf{(Right)} Extracted frontier points in the $n$ vs.\ Compute space. Each circle represents the compute budget $C$ required for a specific $n$ to reach a higher performance bin. The dashed curve shows the fitted scaling law, indicating the optimal $n$ scaling as compute increases.}}
    \label{fig:frontier_tutorial}
\end{figure}

\textbf{Extracting frontiers.} Figure~\ref{fig:frontier_tutorial} provides a schematic illustration of how we extract frontiers and fit the sigmoidal curve.

\section{Additional Compute-Optimal Results}
\label{app:other_results}

\begin{figure}[!ht]
    \centering
    \includegraphics[width=\linewidth]{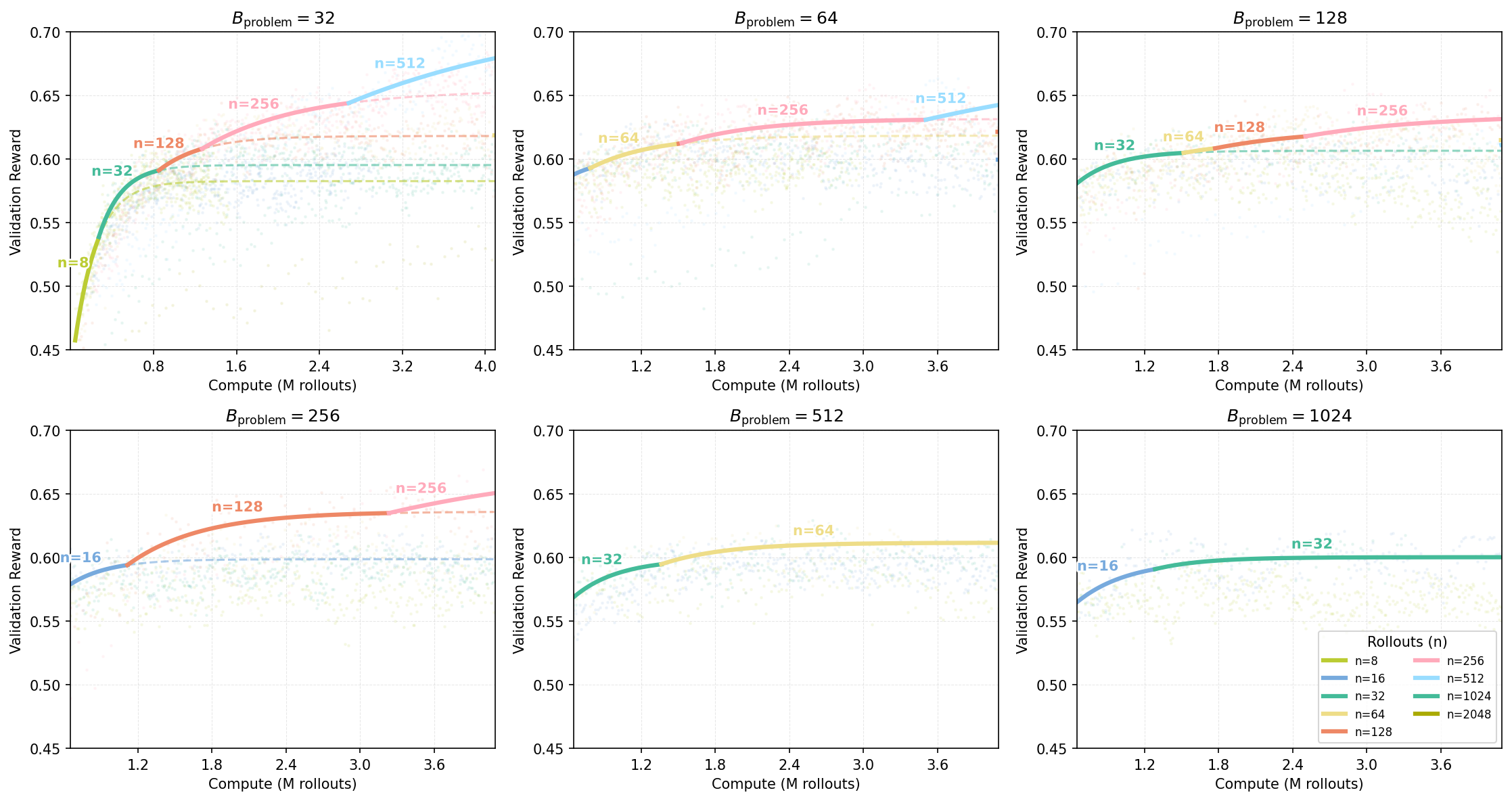}
    \caption{\footnotesize{\textbf{Compute-optimal frontiers maximizing over $n$ varying problems per batch ($B_\text{p}$) on the Easy set.}}}
    \label{fig:appx_fixBprob_easy}
\end{figure}

\begin{figure}[!ht]
    \centering
    \includegraphics[width=\linewidth]{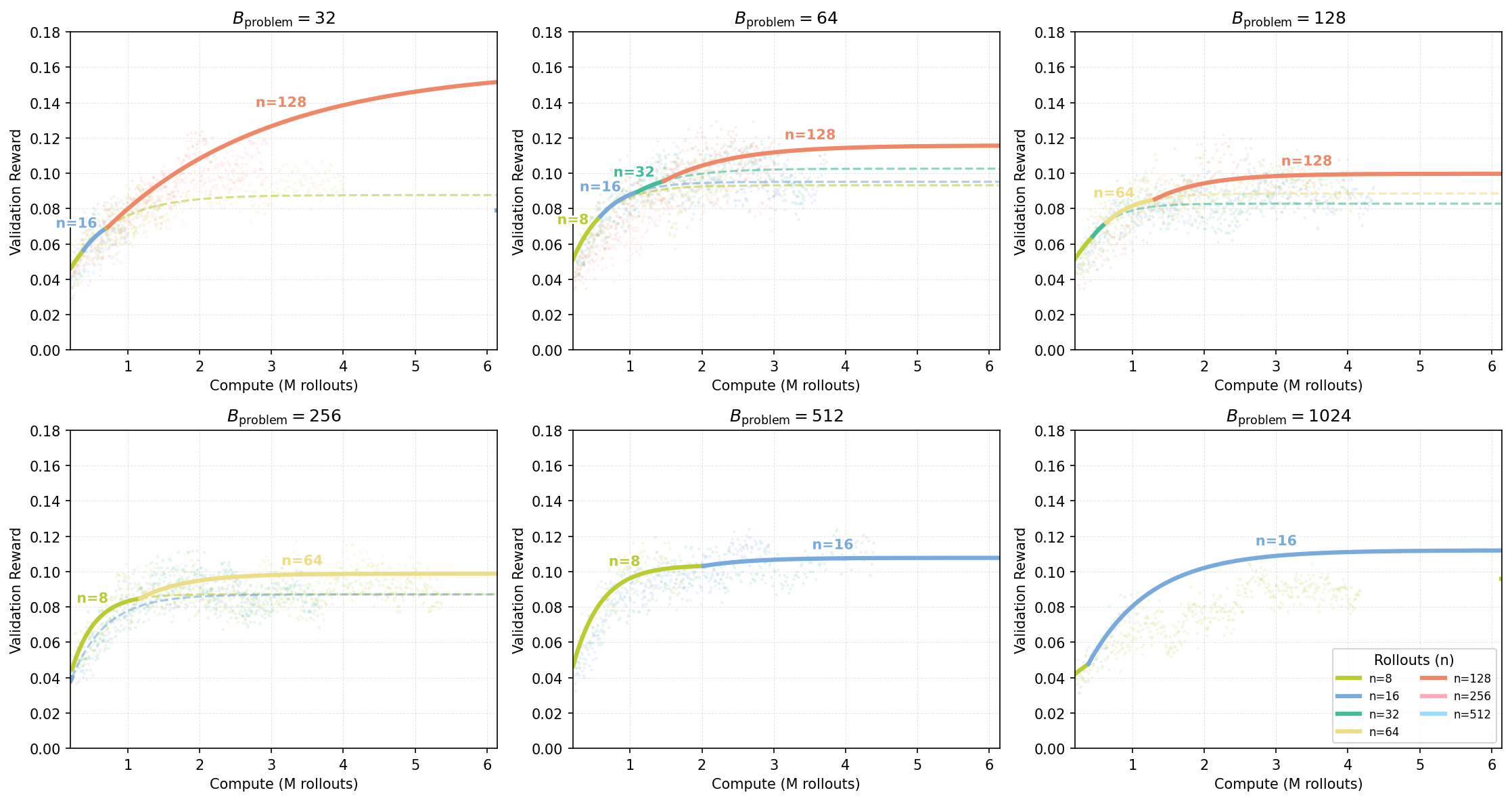}
    \caption{\footnotesize{\textbf{Compute-optimal frontiers maximizing over $n$ varying problems per batch ($B_\text{p}$) on the Hard set.}}}
    \label{fig:appx_fixBprob_hard}
\end{figure}

In the main results, we show one fixed value for $B_\text{p}=32$ for brevity.
Figures~\ref{fig:appx_fixBprob_easy} and~\ref{fig:appx_fixBprob_hard} demonstrate that the scaling trend described in the main text, where larger compute budgets favor increased parallel rollouts ($n$), holds across different fixed values of $B_\text{p}$.
While it appears that larger $B_\text{p}$ settings saturate at lower $n$ values (e.g., $n=16$ at $B_\text{p}=1{,}024$), this might be attributable to the total batch size constraint ($B_{\max} \geq B_\text{p} \cdot n$) in the sweep experiments.
The precise interaction between $B_\text{p}$ and the saturation point of $n$ remains an open question for future investigation.

\begin{figure}[!ht]
    \centering
    \includegraphics[width=\linewidth]{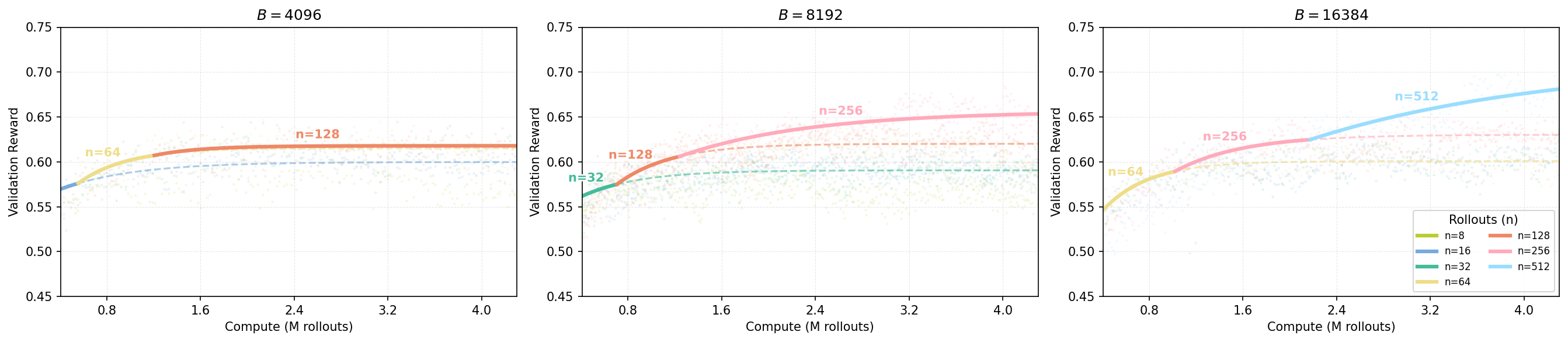}
    \caption{\footnotesize{\textbf{Compute-optimal frontiers on the Easy set under fixed total batch size $B \in \{4096,\ 8192,\ 16384\}$.} Each subplot fixes the total batch size $B$ and sweeps the number of parallel rollouts per problem plotting validation reward versus compute (measured in millions of rollouts).}}
    \label{fig:appx_fixB_easy}
\end{figure}

\begin{figure}[!ht]
    \centering
    \includegraphics[width=\linewidth]{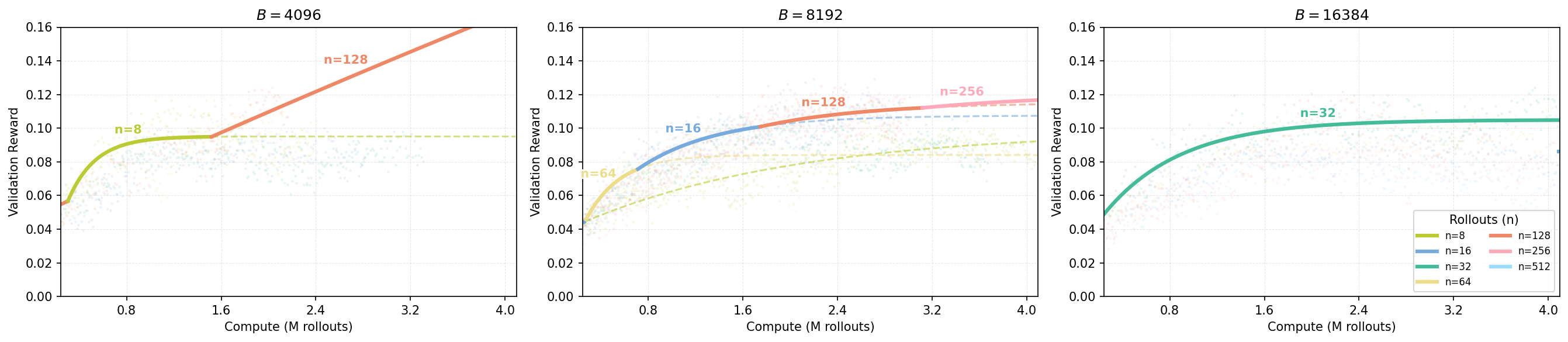}
    \caption{\footnotesize{\textbf{Compute-optimal frontiers on the Hard set under fixed total batch size $B \in \{4096,\ 8192,\ 16384\}$.} Compared to the Easy set, the trends are \textbf{noisier} in the Hard regime. Nevertheless, the qualitative trend remains consistent: as compute increases, the compute-optimal allocation increasingly favors larger parallel rollouts per problem, i.e., larger $n$.}}
    \label{fig:appx_fixB_hard}
\end{figure}

Figure~\ref{fig:appx_fixB_easy} and~\ref{fig:appx_fixB_hard} provide additional compute-optimal frontiers under different fixed values of $B_\text{p}$ on the Easy and Hard splits.
Consistent with Section~3.2, higher sampling budgets increasingly favor larger $n$, indicating that allocating more parallel rollouts per problem is a robust strategy across dataset difficulty and batch-size settings.

\begin{figure}[!ht]
    \centering
    \includegraphics[width=0.9\linewidth]{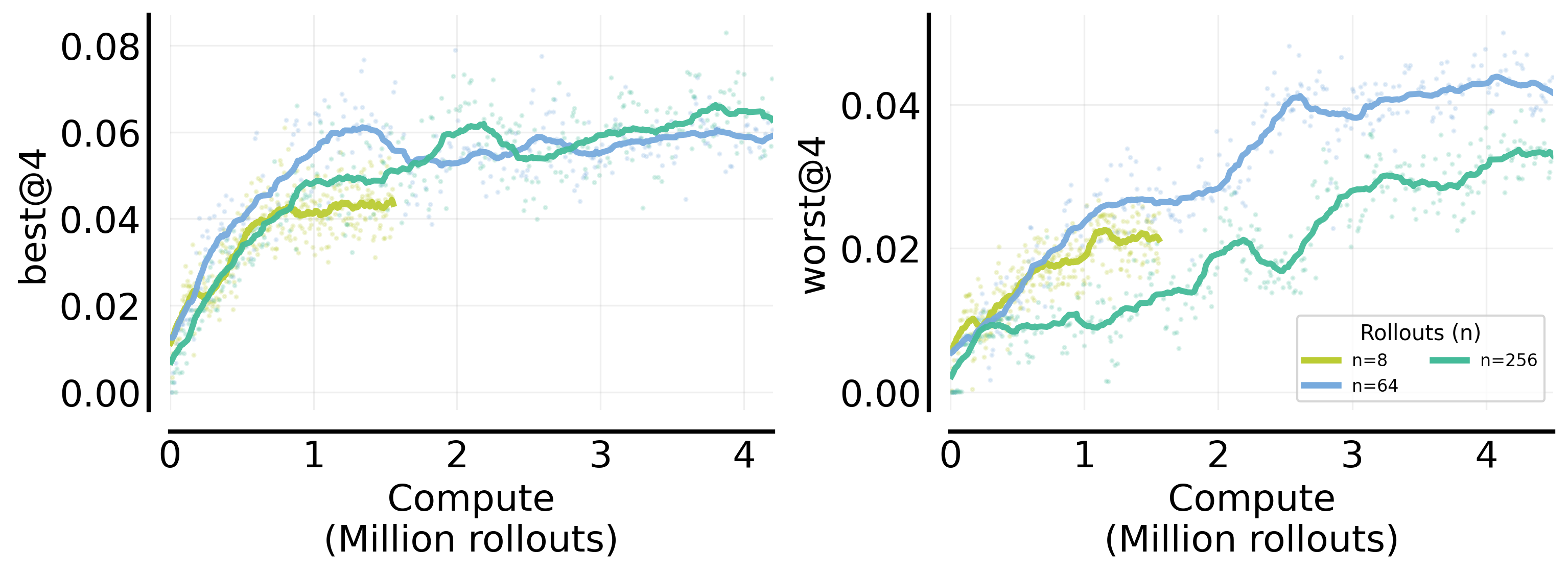}
    \caption{\footnotesize{\textbf{Compute-optimal frontiers on the in-domain Extremely Hard subset (pass@128 $=0$), evaluated with best@4 (left) and worst@4 (right).} Larger $n$ improves \textbf{best@4} at higher compute, whereas \textbf{worst@4} is maximized by a \textbf{moderate} $n=64$, highlighting a strong coverage-sharpening trade-off in the extremely hard regime.}}
    \label{fig:appx_passall0}
\end{figure}

Finally, we report results on the in-domain \emph{Extremely Hard} subset (pass@128 $=0$) using both \textbf{best@4} and \textbf{worst@4} metrics (Figure~\ref{fig:appx_passall0}).
We observe a clear \textbf{coverage--sharpening trade-off}: larger $n$ is more beneficial for improving \textbf{best@4} (coverage), while \textbf{worst@4} (sharpening) is compute-optimally maximized at a \textbf{moderate} $n$ (e.g., $n=64$).
Notably, overly large $n$ (e.g., $n=256$) can underperform on worst@4 despite achieving better coverage, suggesting that the compute-optimal choice of $n$ on extremely hard problems typically lies in an intermediate regime that balances exploration and consistency.

\section{Additional Details: Joint Optimization of $(B_{\text{p}}, n, M)$}
\label{app:q3_results}

In Section~\ref{sec:q3}, we jointly optimize the three sampling axes
$(B_{\text{p}}, n, M)$ under a fixed total rollout compute budget
\begin{equation*}
\label{eq:app_q3_compute}
C \;=\; n \cdot B_{\text{p}} \cdot M .
\end{equation*}
For each compute budget $C$, we exhaustively sweep a grid of feasible pairs $(B_{\text{p}}, n)$ within the range accessible to our system, and set
\begin{equation*}
\label{eq:app_q3_M}
M \;=\; \left\lfloor \frac{C}{n\,B_{\text{p}}} \right\rfloor
\end{equation*}
(up to standard feasibility constraints such as minimum required update steps and hardware throughput limits). We then select the best configuration at each $C$ by
\begin{equation*}
\label{eq:app_q3_argmax}
(B_{\text{p}}^*(C), n^*(C), M^*(C))
\;=\;
\arg\max_{(B_{\text{p}},n,M)\,\in\,\mathcal{G}(C)}
\mathrm{Reward}_{\mathrm{val}}(B_{\text{p}},n,M),
\end{equation*}
where $\mathcal{G}(C)$ denotes the feasible sweep grid at budget $C$ and the validation metric is avg@4 unless stated otherwise.

\begin{figure}[!ht]
    \centering
    \includegraphics[width=0.9\linewidth]{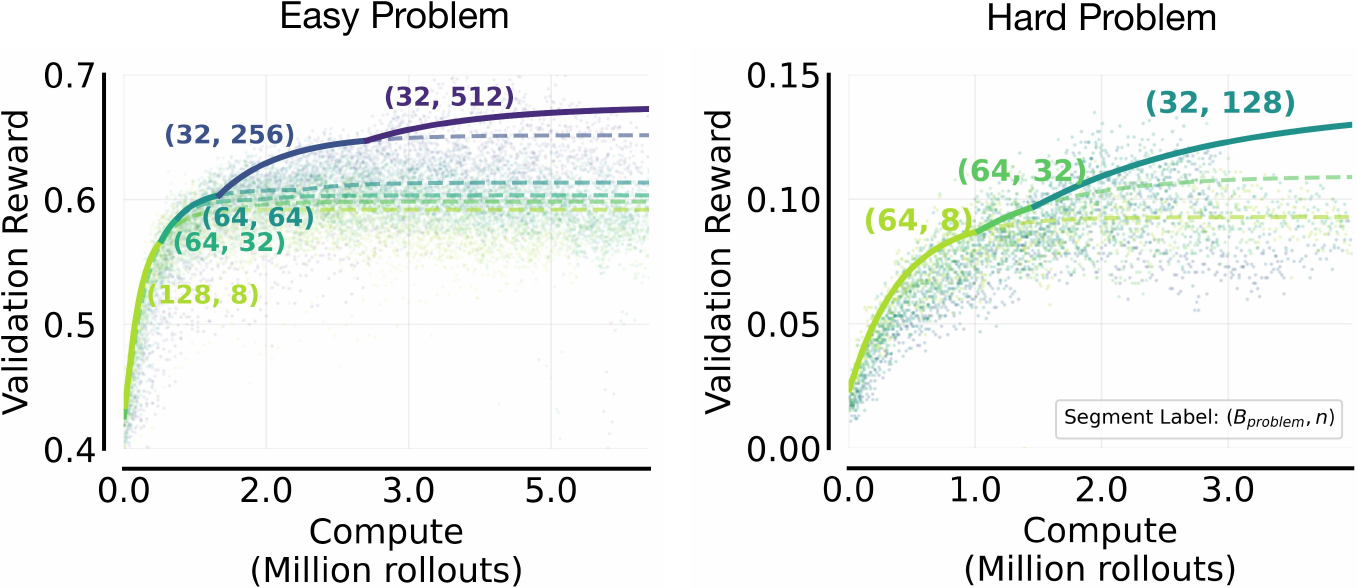}
    \caption{\footnotesize{\textbf{Compute-optimal parallel rollouts $n^*(C)$ under joint optimization of $(B_{\text{p}}, n, M)$.}
    For each total rollout compute budget $C$, we sweep $(B_{\text{p}}, n, M)$ and select the globally best configuration.
    The optimal $n$ increases monotonically with compute and is well-fit by a sigmoid trend on both the Easy \textbf{(left)} and Hard \textbf{(right)} splits.}}
    \label{fig:q3_nstar}
\end{figure}

\begin{figure}[!ht]
    \centering
    \includegraphics[width=0.9\linewidth]{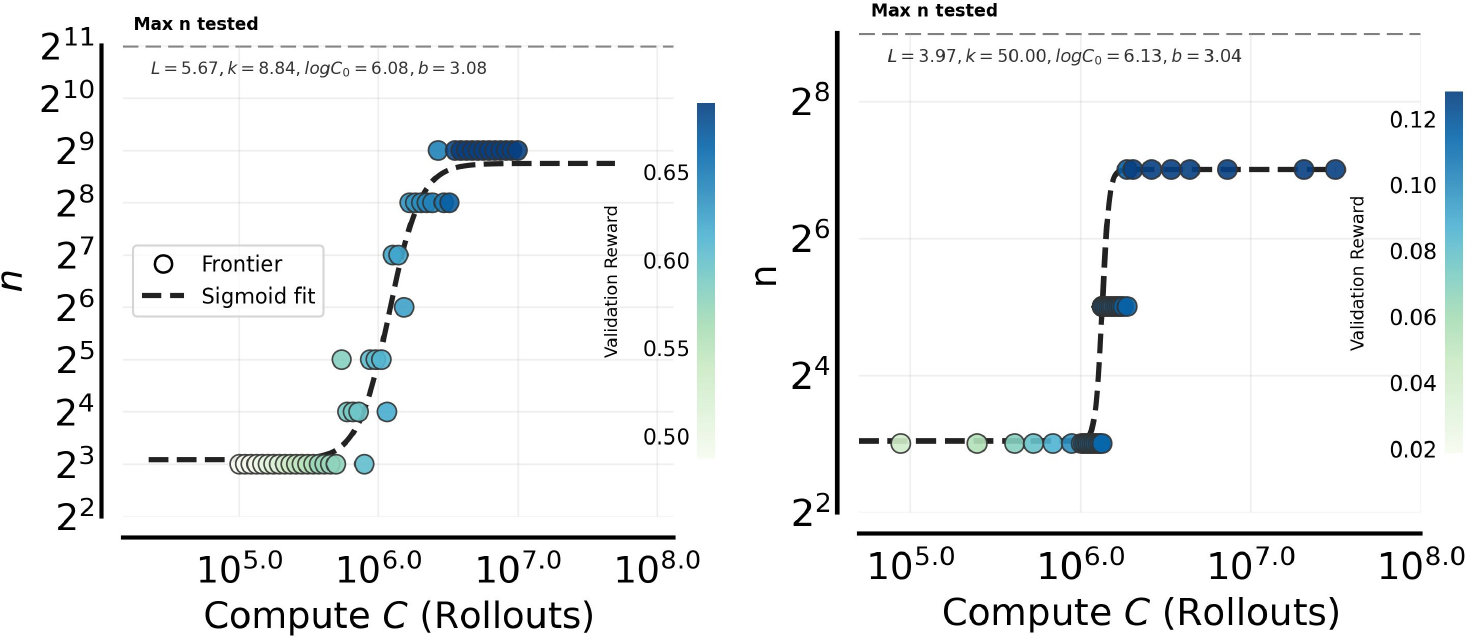}
    \caption{\footnotesize{\textbf{Compute-optimal frontiers from sweeping $(B_{\text{p}}, n, M)$ on Easy and Hard problems.}
    Points on the frontier are annotated by the pre-training sampling configuration $(B_{\text{p}}, n)$, with $M$ determined by the remaining compute.
    Consistent with earlier sections, the frontier shifts to systematically larger $n$ as compute increases.
    In contrast, the frontier-attaining $B_{\text{p}}$ varies across budgets but has only a marginal effect on performance within a moderate range (cf. Section~\ref{sec:q2}).}}
    \label{fig:q3_frontier}
\end{figure}

Across both easy and hard splits, the joint sweep confirms a consistent pattern: the compute-optimal strategy is primarily characterized by the parallel rollouts per problem.
As shown in Fig.~\ref{fig:q3_nstar}--\ref{fig:q3_frontier}, $n^*(C)$ increases monotonically with compute and is well-fit by a sigmoid trend in $\log n$ versus $\log C$.
In contrast, $B_{\text{p}}$ behaves mainly as a \emph{stability constraint} rather than a performance driver:
once $B_{\text{p}}$ is kept within a moderate range, performance varies only weakly with $B_{\text{p}}$, and multiple $B_{\text{p}}$ values can yield similarss results provided training remains stable.
In practice, we therefore recommend the following workflow:
(i) tune $n$ using the fitted $n^*(C)$ curve, (ii) choose the smallest $B_{\text{p}}$ that yields stable training for the target difficulty regime, and (iii) allocate the remaining budget to $M$.

Finally, we note that while our sweeps are exhaustive over the $(B_{\text{p}}, n)$ range we could access, we do not explore regimes with extremely large total rollout sizes where both $B_{\text{p}}$ and $n$ are simultaneously large; understanding interactions at such massive batch sizes is an important direction for future work.

\section{Generalization to OOD tasks}
\label{app:ood_results}

\begin{figure}[!ht]
    \centering
    \includegraphics[width=0.5\linewidth]{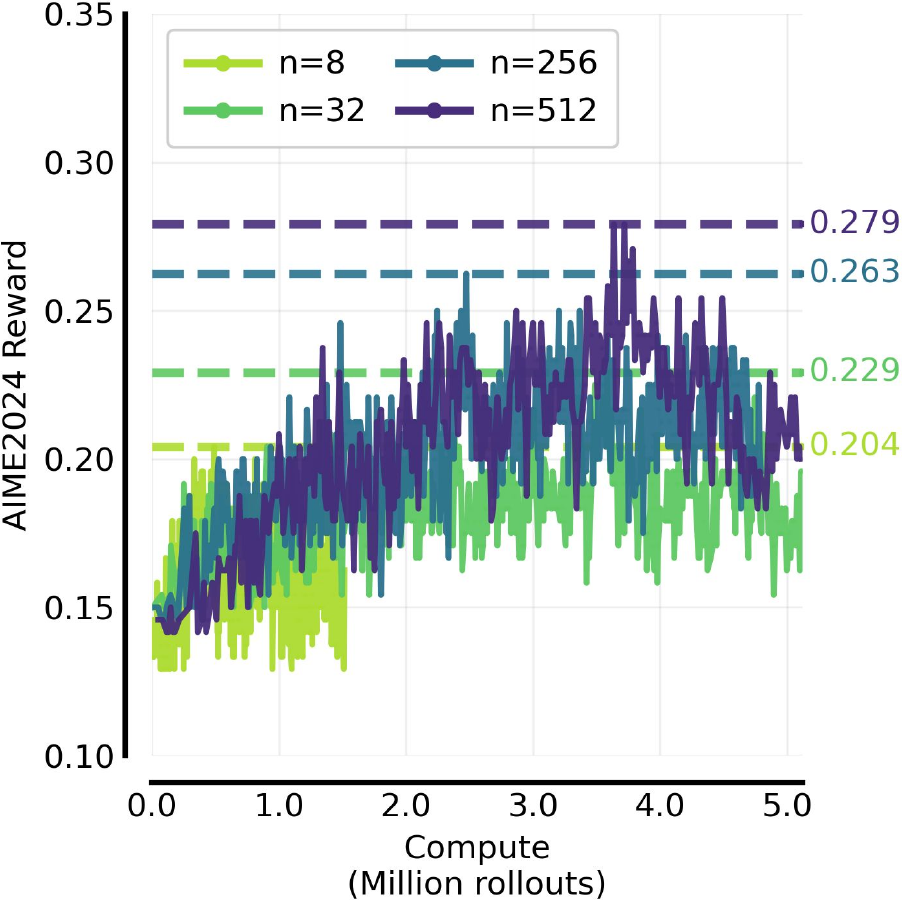}
    \caption{\footnotesize{AIME 24 scores trained with varying parallel rollouts ($n$) under a fixed problem batch size ($B_\text{p}=32$).}}
    \label{fig:appx_aime24}
\end{figure}

In the main text, we prioritize in-domain validation results to minimize the influence of train-test distribution shifts, thereby allowing for a cleaner analysis of compute allocation scaling.
In reality, practical post-training workflows require models to generalize to unseen distributions like downstream tasks.
We examine whether the benefits of increasing parallel rollouts ($n$) extend to out-of-domain (OOD) downstream tasks.
As illustrated in Figure~\ref{fig:appx_aime24}, we observe that larger values of $n$ lead to higher performance on AIME24.

\section{Compute Metrics: Rollouts vs. Tokens}
\label{app:token_view}
\begin{figure}[!ht]
    \centering
    \includegraphics[width=0.9\linewidth]{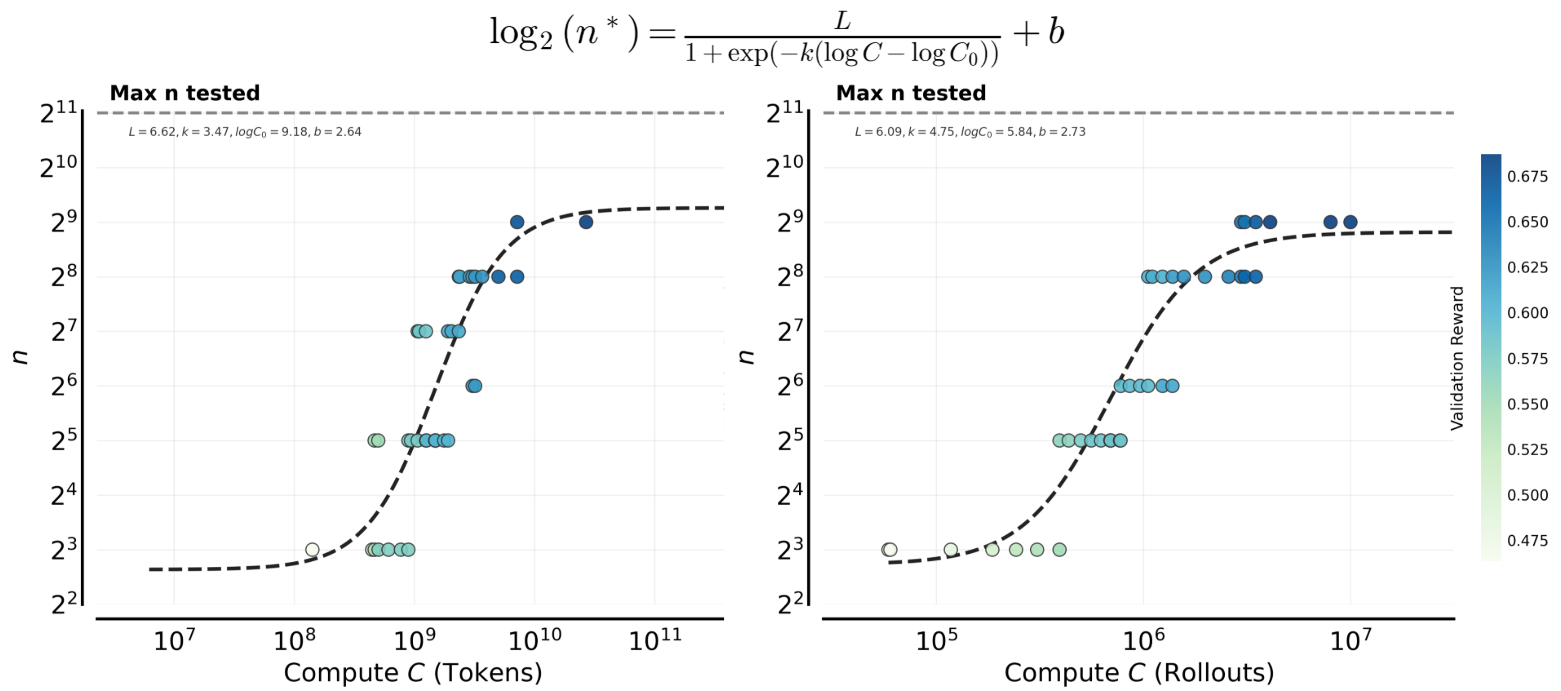}
    \caption{\footnotesize{\textbf{$n^*$ scaling is consistent under token-based vs.\ rollout-based compute.} We fit sigmoid curves for $\log_2(n^*)$ as a function of compute $C$, using either total generated tokens (\textbf{left}) or total rollouts (\textbf{right}). Both choices produce the same qualitative scaling curve---rapid growth followed by saturation---indicating that the compute-optimal $n^*$ trend is robust to the compute definition.}}
    \label{fig:appx_token_compute}
\end{figure}

To verify that our compute--optimal $n^*$ scaling is not an artifact of how we measure compute, we repeat the same fit using another unit: \textbf{total generated tokens}.
As shown in Figure~\ref{fig:appx_token_compute}, both parameterizations lead to an almost identical sigmoid trend.
This suggests that, for our training setup, using rollouts or tokens as the compute proxy makes little practical difference.
The two views are largely related by a near-constant conversion factor governed by the average response length.

One noticeable difference is that the fitted slope parameter $k$ is not exactly the same across the two plots.
This is expected: $k$ controls how sharply $n^*$ transitions as compute increases, and its numerical value depends on the units of $C$.
In experiments, we observe a positive correlation between the model’s response length and validation rewards.
For instance, models at the high-compute frontier tend to have longer response lengths.
Since token-based compute accounts for response length, the $k$ value is smaller, indicating a shallower slope in $n$ scaling relative to compute.
Therefore, the change in $k$ mainly reflects how response length modulates the mapping between rollouts and tokens, rather than a fundamental discrepancy in the underlying scaling behavior.
Nonetheless, the overall scaling trend remains consistent.

\clearpage
\section{Additional Results on Other Algorithms}
\label{app:algos}

For clarity, the main text focuses on the GRPO setting. We also test whether the core compute-allocation insight, that larger parallel rollouts per problem ($n$) become increasingly favorable as total rollout compute grows, especially on harder regimes, extends to other on-policy objectives.
In this appendix, we apply the same $n$-sweep protocol to \textbf{PPO}\cite{schulman2017proximal} and \textbf{CISPO}\cite{minimax2025minimaxm1scalingtesttimecompute}.

We keep the \emph{same} base model (Qwen2.5-7B-Instruct), data splits (Easy/Hard), sampling temperature/top-$p$, and the compute accounting used throughout the paper (compute measured in million rollouts). We sweep $n\in\{16,32,64,128,256\}$ and plot validation reward as a function of compute.
We do not perform an extensive hyperparameter retuning for each algorithm; the goal here is to check whether the qualitative $n$-scaling trend persists beyond GRPO.

\begin{figure}[!ht]
    \centering
    \includegraphics[width=\linewidth]{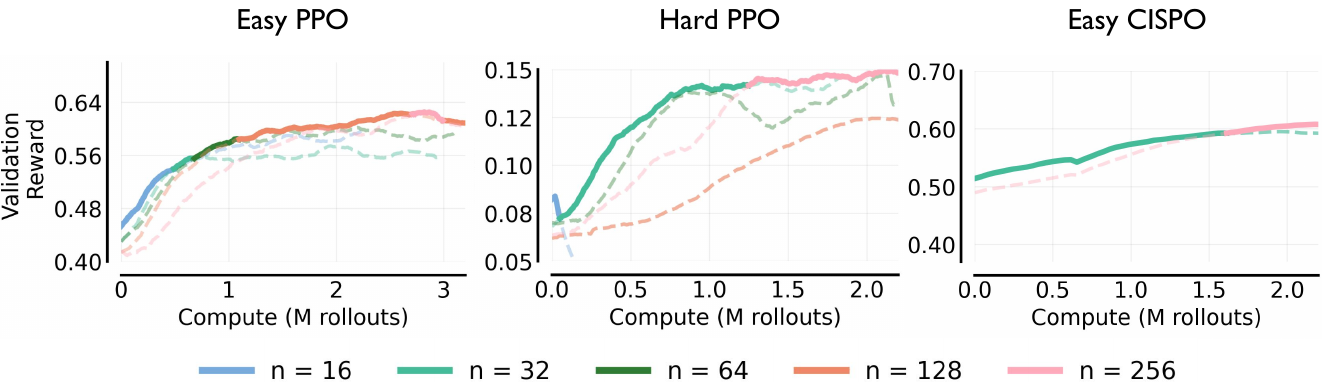}
    \caption{\footnotesize{\textbf{Generalization to other RL algorithms (PPO and CISPO).}
    Validation reward versus compute (million rollouts) for varying parallel rollouts per problem $n$.
    \textbf{Left:} Easy set with PPO. \textbf{Middle:} Hard set with PPO. \textbf{Right:} Easy set with CISPO.
    The qualitative trend matches the main text: as compute increases, larger $n$ becomes increasingly favorable, with a stronger separation on the Hard split.}}
    \label{fig:appx_other_algos}
\end{figure}

Figure~\ref{fig:appx_other_algos} reports reward--compute trajectories under PPO and CISPO.
On \textbf{Hard} with PPO, larger $n$ yields consistently better performance at matched compute, matching the ``discovery-limited'' regime observed in the main text: small $n$ improves slowly while larger $n$ accelerates progress as compute increases.
On \textbf{Easy}, PPO exhibits earlier saturation and weaker separation among large $n$ values, consistent with the Easy regime being less exploration-limited.
CISPO shows a similar qualitative pattern on Easy, with smooth learning curves and competitive performance from moderate-to-large $n$ as compute grows.
Overall, these results suggest that the empirical preference for larger $n$ at higher compute is \emph{not} specific to GRPO's group baseline estimator; it also appears under value-based PPO and an alternative clipped objective (CISPO).

\section{Effects of Reducing Baseline Estimation Variance}
We discuss in the main content how larger $n$ outperforms small $n$ at high compute regimes from exploration and optimization perspectives. Another theoretical advantage of larger $n$ in GRPO is providing a more robust baseline estimator (group average reward), thereby reducing advantage estimate variance. To isolate the gain attributed specifically to precise baseline estimation versus training on more data, we conducted an ablation with a fixed problem batch size ($B_\text{p}=128$). We compared three settings: \textbf{(1) Large $n=256$}, \textbf{(2) Small $n=64$}, and \textbf{(3) Decoupled}, where we generate 256 rollouts to compute high-precision advantage estimates but randomly subsample only 64 rollouts for the policy gradient update.

We observe the best validation reward follows \textbf{(1) $>$ (3) $\approx$ (2)}. The fact that (3) performs similarly to (2) indicates that the benefit of a lower-variance baseline estimator is not significant in this context. Consequently, the superior performance of (1) over (3) suggests that the primary benefit of scaling $n$ stems from broader exploration rather than baseline precision.

\begin{figure}[!ht]
    \centering
    \includegraphics[width=0.75\linewidth]{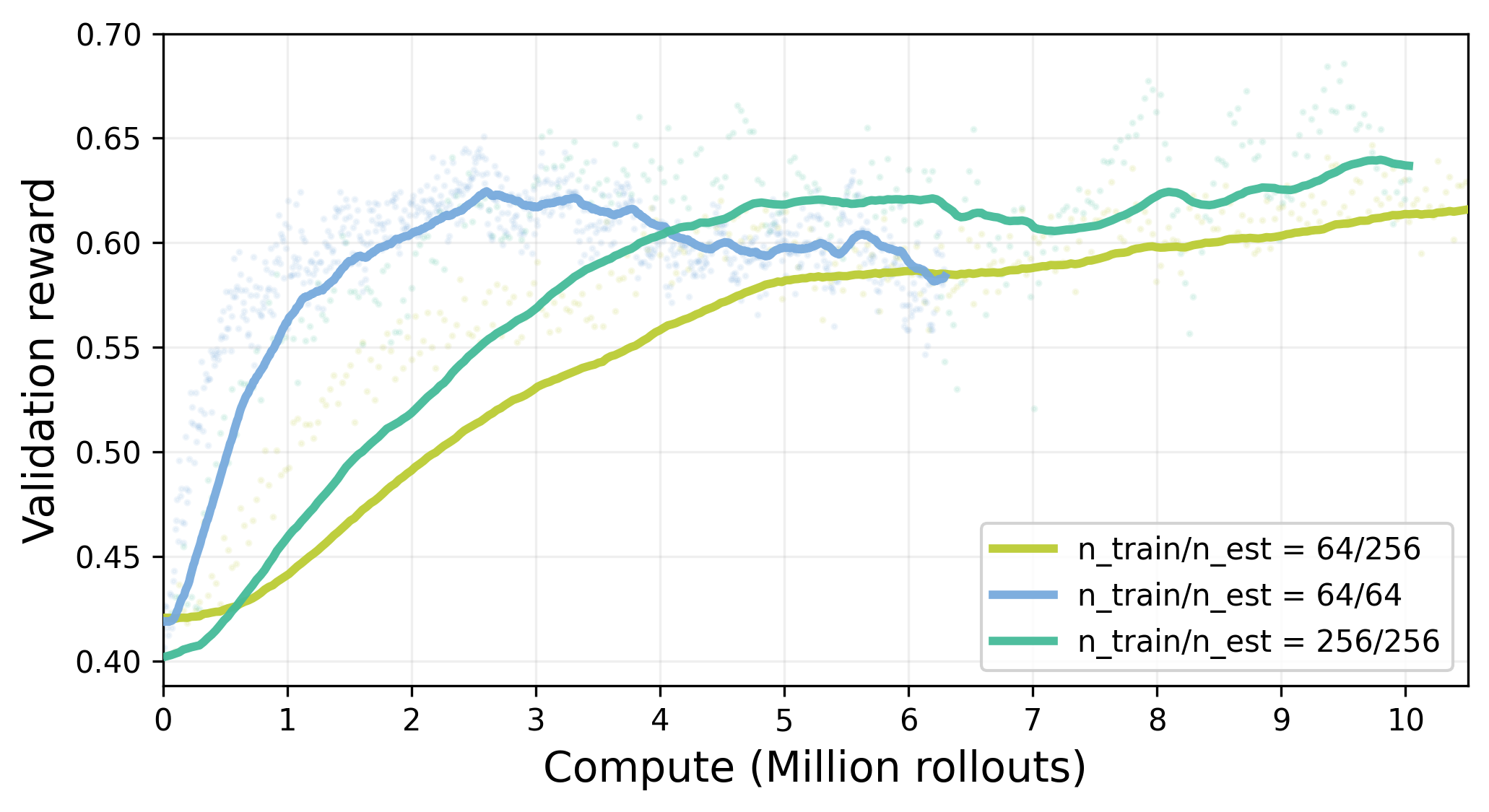}
    \caption{\footnotesize{\textbf{Effects of baseline estimation variance.} Validation reward vs.\ compute (million rollouts) under a fixed problem batch size $B_{\text{p}}=128$, comparing three GRPO settings: (i) large group size $n_{\text{train}}/n_{\text{est}}=256/256$, (ii) small group size $64/64$, and (iii) decoupled baseline estimation $64/256$ (estimate baseline from 256 rollouts but sample 64 from them for the policy-gradient update). We observe consistent ordering \textbf{(1) $>$ (3) $\approx$ (2)}, showing that lower-variance baseline estimation yields negligible gains, while the full $n=256$ run remains best, indicating the dominant gains from scaling $n$ come from broader exploration.}}
    \label{fig:appx_baselineest}
\end{figure}

\section{Base Case: Only One Training Problem}
\label{app:base_case_one_problem}
To build a conceptual model, let us study the simplest setting where we are provided with \emph{one single problem} in the training set. We model this setting as a simple multi-armed bandit problem, where each arm represents one possible response to the problem. We assume training of a tabular softmax policy (i.e., softmax on independently represented logits denoting the response). Please see this for setup~\citep{mei2023stochastic}.

Now let’s say that the base model attains an average pass@1 rate of $p$ on this prompt and say $n$ i.i.d.\ response samples drawn from the policy are used for training at one gradient step. First note that $n$ independent samples change $\text{pass}@n$ exponentially:

\begin{equation*}
    \text{pass@}n = 1-(1-p)^n .
\end{equation*}

Does $n$ change the policy gradient update on the problem in one update? Averaging over $n$ samples does \textbf{not} change the expected policy gradient direction: the expected update is identical to that obtained from a single sample. What it does change is the \textbf{variance} of the gradient estimate, which decreases by a factor $n$.

Prior work~\citep{mei2023stochastic} shows that, when using a single sample per update, tabular (stochastic) softmax policy gradient enjoys an $O(1/t)$ rate on the policy suboptimality (i.e., bound on optimal performance - attained performance) after $t$ update steps. When $n$ independent samples are used by averaging over the policy gradient update, repeating the same analysis yields
\begin{equation*}
    \mathbb{E}\Big[\text{suboptimality at step }t\Big]
= O\left(\frac{A}{n\cdot t} + \frac{B}{t}\right),
\end{equation*}

where $B \ll A$ is a constant that does not depend on the variance of the policy gradient estimate. The constant $A$ in $\frac{A}{n\cdot t}$ depends on variance in the policy gradient estimate and corresponds to the leading term (for reasonably small $n$).

With this guarantee, the convergence rate is still linear in $t$, but the effect of stochasticity reduces drastically. For the term $\frac{A}{n\cdot t}$, $n$ and $t$ can be interchanged: one can reduce the error in this term by using a larger $n$ for a smaller $t$. The other term depends only on $t$, indicating that out of all compute allocation configurations in Section~\ref{sec:q1}, \textbf{for instance, one should prefer the configuration that makes more sequential updates $M$ as opposed to choosing a larger $n$.} However, this is not the case in practice.

\section{A Mental Model for Interference}
\begin{figure}[!ht]
    \centering
    \includegraphics[width=\linewidth]{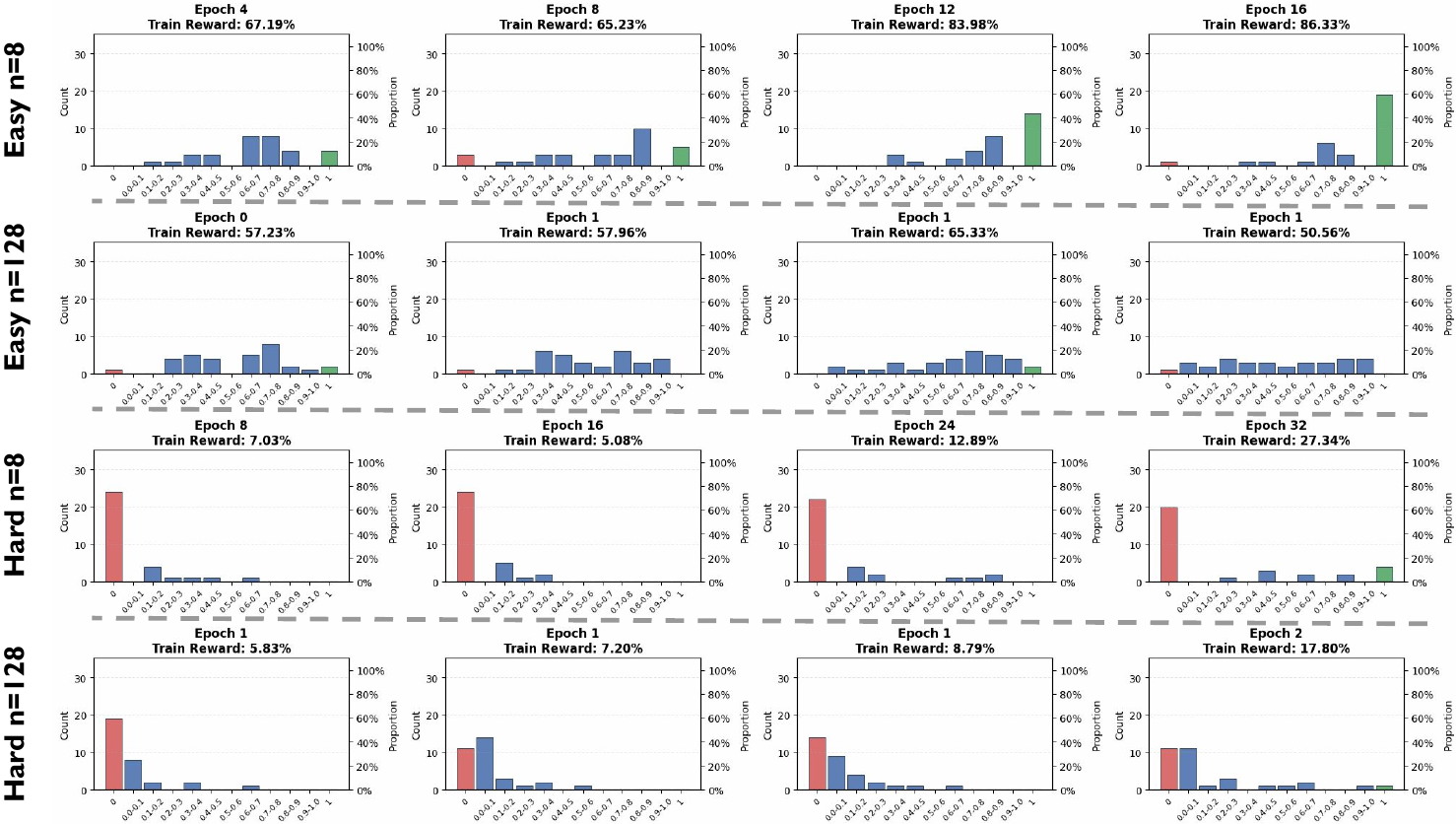}
    \caption{\footnotesize{\textbf{Dynamics of pass@1 distributions (sanity-checking the interference analysis in Fig.~\ref{fig:train_pass1_dist}).}
    We visualize the evolution of pass@1 histograms across training for the same four cases (Easy/Hard $\cdot$ $n\!=\!8/128$) at matched compute.
    The temporal trajectories corroborate the main-text interpretation: on \textbf{Easy}, small $n$ progressively polarizes into a mass near 1 with a persistent non-zero fraction near 0 (optimization-induced \emph{interference}), whereas large $n$ maintains a more dispersed, uniform distribution.
    On \textbf{Hard}, large $n$ increases \emph{coverage} by steadily reducing the zero-mass, while small $n$ concentrates gains on a subset of solvable problems, yielding sharper but less comprehensive improvements.}}
    \label{fig:appx_dynamics_dist}
\end{figure}

A natural diagnostic is the distribution of $\text{pass}@1$ across prompts. Inference-time scaling laws~\cite{arxiv-org-2502-17578} relate $\text{pass}@n$ to the population $\text{pass}@1$ distribution, but RL training differs because the model learns from the $n$ rollouts it produces, and updates across problems introduce interference. A useful mental model is that interference is smaller when learning progress is distributed roughly uniformly across prompts. Thus, in the Fig. ~\ref{fig:appx_dynamics_dist}, changes in the $\text{pass}@1$ distribution over training can serve as a diagnostic: uniform improvement suggests controlled interference, while highly uneven improvement suggests strong interference and rich-gets-richer dynamics.

\end{document}